\documentclass{article} 
\usepackage[final]{colm2026_conference}
\usepackage{graphicx}
\usepackage{microtype}
\usepackage{hyperref}
\usepackage{url}
\usepackage{booktabs}
\usepackage{amsmath}
\usepackage{longtable}
\usepackage{xspace}
\usepackage{wrapfig}

\usepackage{lineno}
\usepackage{float} 
\definecolor{darkblue}{rgb}{0, 0, 0.5}
\hypersetup{colorlinks=true, citecolor=darkblue, linkcolor=darkblue, urlcolor=darkblue}

\title{EvoLen: Evolution-Guided Tokenization for DNA Language Model}

\author{
\textbf{Nan Huang}$^{1}$, \enspace
\textbf{Xiaoxiao Zhou}$^{2}$, \enspace
\textbf{Junxia Cui}$^{1}$, \enspace
\textbf{Mario Tapia-Pacheco}$^{1}$, \\
\textbf{Tiffany Amariuta}$^{1}$, \enspace
\textbf{Yang E. Li}$^{2,\ast}$, \enspace
\textbf{Jingbo Shang}$^{1,\ast}$ \\
\vspace{3pt}
$^{1}$University of California, San Diego, La Jolla, CA 92093, USA \\
$^{2}$Washington University in St.\ Louis, St.\ Louis, MO 63130, USA \\
\vspace{3pt}
{\small\texttt{\{n5huang, jucui, mtapiapacheco, tamariuta, jshang\}@ucsd.edu}} \\
{\small\texttt{\{xiaoxiao.zhou, yeli\}@wustl.edu}} \\
\vspace{3pt}
$^{\ast}$Co-corresponding authors.
}

\begin{document}

\ifcolmsubmission
\linenumbers
\fi

\maketitle

\begin{abstract}

Tokens serve as the basic units of representation in DNA language models (DNALMs), yet their design remains under-explored. Unlike natural language, DNA lacks inherent token boundaries or predefined compositional rules, making tokenization a fundamental modeling decision rather than a naturally specified one. While existing approaches like byte-pair encoding (BPE) excel at capturing token structures that reflect human-generated linguistic regularities, DNA is organized by biological function and evolutionary constraint rather than linguistic convention. We argue that DNA tokenization should prioritize functional sequence patterns like regulatory motifs—short, recurring segments under evolutionary constraint and typically preserved across species. We incorporate evolutionary information directly into the tokenization process through \textbf{EvoLen}, a tokenizer that combines \textbf{Evo}lutionary stratification with \textbf{Len}gth-aware decoding to better preserve motif-scale functional sequence units. EvoLen uses cross-species evolutionary signals to group DNA sequences, trains separate BPE tokenizers on each group, merges the resulting vocabularies via a rule prioritizing preserved patterns, and applies length-aware decoding with dynamic programming. Through controlled experiments, EvoLen improves the preservation of functional sequence patterns, differentiation across genomic contexts, and alignment with evolutionary constraint, while matching or outperforming standard BPE across diverse DNALM benchmarks. These results demonstrate that tokenization introduces a critical inductive bias and that incorporating evolutionary information yields more biologically meaningful and interpretable sequence representations. Code, pretrained and fine-tuned checkpoints, and tokenizer files are available at
\url{https://github.com/HN020719/EvoLen} and \url{https://huggingface.co/EvoLenTokenizer}.

\end{abstract}

\vspace{-2mm}
\section{Introduction}

\begin{figure*}[t]
    \centering
    \includegraphics[width=\textwidth]{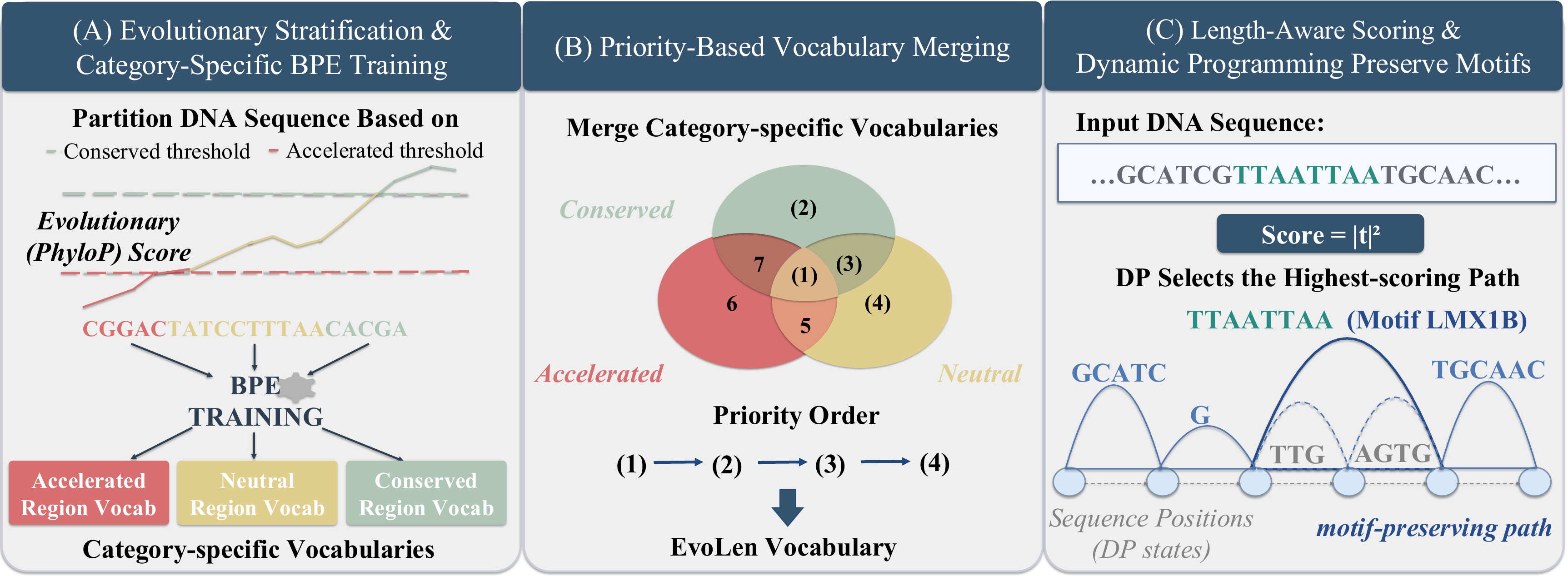}
    \caption{\textbf{EvoLen tokenization pipeline.}
    \textbf{A)} Partition the genome into conserved, neutral, and accelerated regions using phyloP scores and train a category-specific BPE tokenizer on each sequence pool.
    \textbf{B)} Merge the resulting vocabularies using a conservation-prioritized rule.
    \textbf{C)} Apply length-aware scoring to the merged vocabulary and dynamic programming to ensure optimal, non-overlapping segmentation. This enables EvoLen to preserve more functional motifs as intact tokens than BPE, enhancing representation coherence.}
    \label{fig:pipeline}
\end{figure*}

DNA language models (DNALMs) require transforming continuous nucleotide sequences into discrete tokens. Unlike natural language, DNA does not come with intrinsic boundaries analogous to words, so tokenization is not a neutral preprocessing step. Instead, it determines the units through which sequence structure is exposed to the model, and therefore shapes what patterns can be represented efficiently. Most existing genomic tokenization strategies inherit NLP paradigms: fixed-length $k$-mers \citep{ji2021dnabert} impose uniform scales, while adaptive methods like byte-pair encoding (BPE) \citep{sennrich2016neural}, applied to DNA \citep{zhou2023dnabert2},  optimize for frequency and compression. However, these objectives often ignore biological function, leading to the fragmentation of coherent functional units \citep{provilkov2019bpe}.

This mismatch is especially important in regulatory genomics, where many predictive signals are carried by short functional sequence patterns. A key example is transcription factor (TF) binding motifs---short DNA segments (typically 6--12 bp) recognized by regulatory proteins that control gene expression \citep{avsec2021bpnet}. When BPE splits such a pattern into multiple pieces or merges part of it with flanking sequence, the intact functional unit is no longer explicitly represented, making the full regulatory pattern harder for the model to learn. As illustrated by the examples in Appendix Figure~\ref{fig:motif_fragmentation_appendix}, standard BPE breaks the LMX1B binding sequence \texttt{TAATTAA} into shorter substrings and merges part of the motif with neighboring bases. These limitations suggest that genomic tokenization should account for biological structure rather than substring frequency alone. 

One broadly available signal for such structure is evolutionary constraint: functional genomic elements are more likely to be preserved across species, whereas less constrained or lineage-specific regions may evolve more freely. Several signals could in principle capture this structure, including chromatin accessibility, TF binding annotations, or comparative genomics-based conservation scores such as phyloP \citep{siepel2010evolutionary, ucsc_phylop}, phastCons \citep{siepel2005phastcons}, and GERP \citep{davydov2010gerp}. Among these, phyloP is particularly well suited to tokenizer construction because it is available at single-nucleotide resolution, provides signed scores that distinguish conserved from accelerated positions, and is derived from cross-species sequence comparison rather than from any downstream task label. This makes it a genome-wide, fine-grained, and task-agnostic signal for identifying sequence regions likely to carry functional information.

This motivates a central question: \textit{what properties should tokens satisfy to better capture regulatory sequence structure?} To address this, we formalize these desired properties in Section~2 and introduce \textbf{EvoLen}, a tokenizer that integrates \textbf{Evo}lutionary stratification with \textbf{Len}gth-aware decoding through a \textbf{partition-and-merge} strategy. As illustrated in Figure~\ref{fig:pipeline}, EvoLen first \textbf{partitions} the genome into three sequence pools based on phyloP scores---representing highly conserved regions, where substitutions are rare across species and which typically carry function; weakly constrained (neutral) regions, which evolve at the background rate; and rapidly evolving (accelerated) regions, which change faster than the background. It then trains separate BPE tokenizers on each pool and \textbf{merges} the resulting vocabularies into a single set using a priority rule that favors tokens from evolutionarily preserved regions. Finally, EvoLen applies length-aware scoring to the merged vocabulary and dynamic programming to ensure optimal, non-overlapping sequence segmentation. 

Both token structural evaluations and downstream benchmarks confirm that \textbf{EvoLen} preserves functional sequence units while maintaining a flexible subword vocabulary, providing a critical inductive bias for genomic modeling. Our main contributions are as follows:

\begin{enumerate}
    \item \textbf{EvoLen Tokenizer:} We introduce a conservation-aware tokenizer that incorporates evolutionary information into vocabulary construction via a \textbf{partition-and-merge} strategy and utilizes length-aware decoding for optimal segmentation.
    
    \item \textbf{Biological Interpretability:} We demonstrate that \textbf{EvoLen} yields more biologically meaningful representations. At a vocabulary size of $5{,}120$, it preserves transcription factor (TF) motifs as single tokens $27.4\%$ more frequently than standard BPE. Furthermore, it enhances regulatory specificity, increasing the Jensen--Shannon distance between promoter and enhancer token-length distributions by $47\%$ , and improves alignment with evolutionary signals by up to $14.4\%$.
    
    \item \textbf{Downstream Performance:} We show that informed tokenization alone improves genomic modeling without altering model architecture or pre-training objectives. Despite using a smaller pretraining budget, \textbf{EvoLen} outperforms the baseline on 11 of 15 task groups and ranks in the top two among four tokenization strategies on 13 groups, including TF binding and cross-species brain cell-type prediction.
\end{enumerate}

\vspace{-2mm}
\section{Problem Setup}

Given a DNA sequence $S = (s_1, s_2, \dots, s_n)$, a tokenizer maps $S$ to a sequence of contiguous tokens $(t_1, t_2, \dots, t_k)$. This segmentation defines the basic units available to the model and therefore determines the scale at which sequence patterns can be represented and composed. For genomic sequence modeling, our goal is to construct tokens that better align with biological features; the four structural properties below define the criteria that guide the EvoLen pipeline in Figure~\ref{fig:pipeline} and the evaluations in Section~4. In particular, desirable genomic tokens should satisfy: \textbf{(P1) Functional Integrity}: preserve short functional units such as transcription factor binding motifs; \textbf{(P2) Regulatory Specificity}: produce tokenization patterns that differ across genomic contexts such as promoters and enhancers; \textbf{(P3) Evolutionary Consistency}: align token boundaries with regions under similar evolutionary constraint; \textbf{(P4) Pattern Recurrence}: capture recurrent functional sequence patterns rather than merely frequent substrings. Our goal is therefore to construct a tokenizer whose vocabulary construction and decoding procedure better satisfy these properties than standard frequency-driven tokenization.

\vspace{-2mm}
\section{EvoLen Tokenizer Construction}
We now describe how EvoLen is constructed. The tokenizer consists of three components corresponding to the pipeline in Figure~\ref{fig:pipeline}: evolutionary stratification with category-specific BPE training, conservation-prioritized vocabulary merging, and length-aware decoding via dynamic programming.

\subsection{Evolutionary Stratification and Category-Specific BPE}

\paragraph{Why stratify the genome?}
Different parts of the genome are shaped by different evolutionary pressures. Conserved regions are more likely to contain functionally constrained elements, whereas neutral and accelerated regions reflect different sequence dynamics. If all genomic regions are pooled together during token discovery, these signals can be blurred, making it harder for the tokenizer to recover biologically coherent units.

\paragraph{How we stratify.}
We therefore use phyloP scores \citep{ucsc_phylop}, which quantify deviations in substitution rate across species, to partition the human genome \citep{ucsc_genome_browser} into three evolutionary categories: conserved ($\mathrm{con}$), neutral ($\mathrm{neu}$), and accelerated ($\mathrm{acc}$). To obtain stable regional signals, we first divide the genome into non-overlapping 100\,bp bins and compute the mean phyloP score within each bin. Let $x_b$ denote the mean phyloP score of bin $b$. We then compute the global mean $\mu$ and standard deviation $\sigma$ over all bins, and assign each bin using a two-tailed Z-score rule:
\[
\text{conserved: } x_b > \mu + z\sigma, \qquad
\text{accelerated: } x_b < \mu - z\sigma, \qquad
\text{neutral: } \mu - z\sigma \le x_b \le \mu + z\sigma,
\]
where $z = 1.645$, corresponding to a two-tailed significance level of $p < 0.1$. Thus, bins with significantly positive phyloP values are assigned to the conserved category, bins with significantly negative values are assigned to the accelerated category, and the remaining bins are treated as neutral.

This procedure yields three sequence pools with distinct evolutionary profiles. Conserved bins are enriched for regions under purifying selection, accelerated bins capture regions with elevated substitution rates, and neutral bins provide a broad genomic background.

\paragraph{Category-specific BPE training.}
We then independently train a BPE tokenizer on each of the three sequence pools to learn a candidate vocabulary for that evolutionary regime. This produces three category-specific vocabularies that capture distinct merge patterns, rather than forcing a single tokenizer to absorb all sequence contexts at once.

\subsection{Priority-Based Vocabulary Merging}

\paragraph{Why build vocabularies separately first?}
After stratification, each evolutionary category contains a different mix of sequence patterns. Conserved regions are more likely to contain reusable functional units, whereas neutral and accelerated regions contribute broader background diversity. Training a single tokenizer across all regions can blur these differences. We therefore first learn candidate vocabularies within each category and then merge them using a conservation-prioritized rule that favors reusable sequence structure.
\paragraph{How we merge them.}
For each category $c \in \{\text{con}, \text{neu}, \text{acc}\}$, we learn a candidate vocabulary $V_c$. We then construct the final vocabulary using the priority order illustrated in Figure~\ref{fig:pipeline}B: (1) tokens shared across all three regions; (2) conserved-specific tokens; (3) tokens shared by conserved and neutral regions but not accelerated regions; and (4) neutral-specific tokens to fill the remaining capacity.

This rule biases the final vocabulary toward sequence units that are stable across functionally constrained genomic contexts while still retaining sufficient coverage of the broader genome. In practice, it favors tokens that are more likely to reflect biologically reusable structure rather than purely local frequency artifacts.

\subsection{Length-Aware Scoring and Dynamic Programming Decoding}
\paragraph{Why is an additional decoding step needed?}
Even after conservation-aware vocabulary construction, segmentation can still over-prefer short, frequent substrings. A biologically coherent motif may already exist in the vocabulary, yet a standard frequency-based decoder can still split it into shorter pieces if those pieces receive better local scores. This creates a mismatch between what enters the vocabulary and what is actually used during tokenization. To reduce this unnecessary fragmentation and better preserve functional integrity, EvoLen adds an explicit length-aware preference during decoding.

\paragraph{Scoring rule.}
After vocabulary construction, the merged token set is serialized as a Unigram tokenizer \citep{kudo2018subword, kudo2018sentencepiece}. Each token $t$ is assigned a score proportional to the square of its length, $\mathrm{score}(t)=|t|^2$, where $|t|$ denotes the number of nucleotides in token $t$. This scoring rule does not force long tokens everywhere. Instead, it rewards longer units when they are already supported by the vocabulary, while still allowing short tokens when no coherent multi-base pattern is available.

\paragraph{Dynamic programming decoding.}
Given a DNA sequence $s = s_1 s_2 \cdots s_n$ and a scored vocabulary $V$, we use dynamic programming to select the globally optimal non-overlapping segmentation of $s$. For a sequence prefix ending at position $i$, let $\mathrm{DP}[i]$ denote the maximum achievable total score for the prefix $s_{1:i}$, with $\mathrm{DP}[0]=0$. We then compute
$\mathrm{DP}[i] = \max_{j < i,\; s_{j+1:i} \in V} \left( \mathrm{DP}[j] + |s_{j+1:i}|^2 \right)$. This guarantees that the final tokenization is globally optimal under the length-aware objective. 

By construction, the decoder prefers longer intact tokens over fragmented alternatives spanning the same span, which in practice reduces fragmentation into short high-frequency substrings and helps preserve motif-scale functional units. This places EvoLen's decoder in the same family as shortest-path tokenizers: UnigramLM performs Viterbi decoding over a scored vocabulary \citep{kudo2018subword}, and PathPiece makes the shortest-path formulation explicit \citep{schmidt2024tokenization}. EvoLen differs in its scoring function, which rewards length rather than corpus likelihood.

Together, this three-stage pipeline better preserves motif boundaries and yields more coherent sequence representations, as illustrated by the intact motif examples in Appendix Figure~\ref{fig:motif_fragmentation_appendix}.

\vspace{-2mm}
\section{Functional and Evolutionary Token Analysis}
We next test whether EvoLen satisfies the four properties defined in Section~2. Specifically, we ask whether it better preserves short functional sequence units (P1), produces regulatory element-specific tokenization patterns (P2), aligns with evolutionary constraint (P3), and captures recurrent functional sequence structure (P4). P1, P2 and P4 are evaluated against signals held out from construction, including JASPAR motifs, promoter/enhancer/exon/intron annotations, and a neutral intronic background. None of these are seen by the tokenizer during vocabulary learning. P3 uses phyloP, which does enter construction, but only at the level of 100~bp pools, whereas the evaluation measures per-token conservation, a granularity neither BPE merging nor length-aware decoding optimizes directly.

\subsection{Motif Preservation}

\paragraph{Why this matters.} Many regulatory functions in DNA are mediated by short sequence patterns called TF motifs \citep{alipanahi2015predicting, kelley2018sequential}. A tokenizer that preserves such a motif as one token presents it to the model as a single coherent unit; a tokenizer that splits it forces the model to reconstruct that signal across multiple tokens. We therefore evaluate whether EvoLen better preserves intact motif-scale units, corresponding to \textbf{(P1) Functional Integrity}.

\paragraph{How we evaluate it.} For each vocabulary size, we compute the \textbf{perfect match rate}: the fraction of known motifs that are encoded as a single token without being split. We use transcription factor motifs from the JASPAR 2024 vertebrate library \citep{bailey2009meme, grant2011fimo}, represented as PWMs, and convert them into fixed motif sequences by thresholding each position at 0.5, trimming wildcard positions at both ends, and retaining the highest-probability nucleotide at the remaining positions. We restrict the analysis to motifs of length at most 12 bp \citep{dnamotiftokenizer2025}. This directly measures how often the tokenizer preserves a complete functional sequence element as one representational unit. Additional motif tokenization diagnostics, including fragmentation, coverage, and consistency across motif variants, are reported in Appendix Figure~\ref{fig:motif_preservation_appendix}.

\paragraph{Main result.} EvoLen achieves a higher perfect match rate than baseline BPE at every vocabulary size, with relative gains of $+9.1\%$ at 2,048, $+3.8\%$ at 3,072, $+17.5\%$ at 4,096, and $+27.4\%$ at 5,120 (Figure~\ref{fig:structural_analysis})A. The largest improvement appears at the vocabulary size used for downstream evaluation, where EvoLen preserves substantially more motifs as intact tokens. Overall, these results support \textbf{(P1) Functional Integrity}: EvoLen more often represents short functional DNA elements as single units rather than fragmented substrings.

\vspace{-2mm}
\subsection{Functional Region Token-Length Signatures}

\paragraph{Why this matters.} Different regulatory elements play different roles. Promoters help initiate gene expression \citep{seizl2011ga_element}, enhancers help regulate when and where genes are activated, and exons encode the transcribed sequence itself. If a tokenizer captures the feature of these biological structures, it should not segment all of these regions in the same way. We therefore test whether EvoLen produces more distinct tokenization patterns across various regulatory elements, corresponding to \textbf{(P2) Regulatory Specificity}.

\begin{figure*}[t]
\centering
\begin{minipage}[t]{0.32\textwidth}
    \centering
    \includegraphics[width=\linewidth, trim=0 5 0 5, clip]{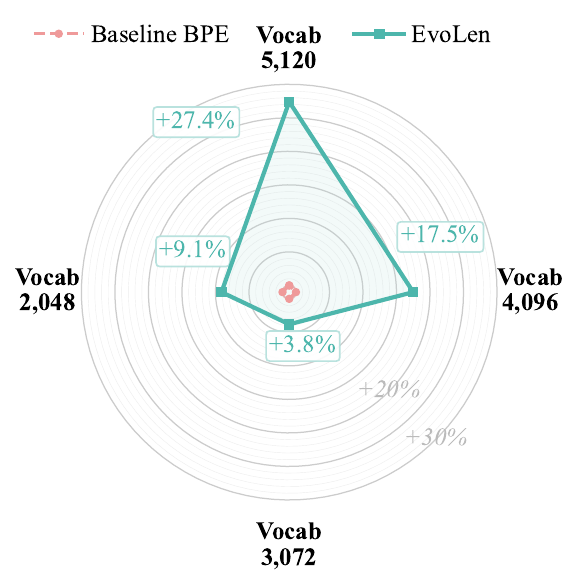}
\end{minipage}\hspace{1mm}
\begin{minipage}[t]{0.24\textwidth}
    \centering
    \includegraphics[width=\linewidth, trim=0 5 0 5, clip]{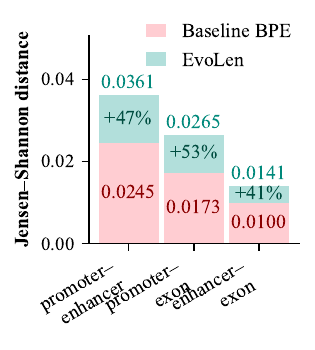}
\end{minipage}\hspace{1mm}
\begin{minipage}[t]{0.41\textwidth}
    \centering
    \includegraphics[width=\linewidth, trim=0 5 0 5, clip]{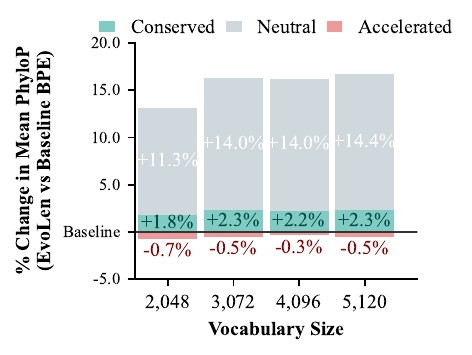}
\end{minipage}
\vspace{-2mm}
\caption{\textbf{(A)} EvoLen increases the fraction of motifs preserved as single tokens across all vocabulary sizes (P1). \textbf{(B)} Increased Jensen–Shannon distances at vocabulary size 5,120 demonstrate that EvoLen produces more distinct token-length distributions between promoters, enhancers, and exons than BPE (P2).  \textbf{(C)} Relative gains in mean phyloP scores signify enhanced alignment with evolutionary conservation (P3).}

\label{fig:structural_analysis}
\vspace{-4mm}
\end{figure*}

\paragraph{How we evaluate it.} At vocabulary size 5,120, we compare token-length distributions in promoters, enhancers, and exons. Token lengths are grouped into four bins (1--2 bp, 3--5 bp, 6--8 bp, and 9+ bp), and we measure the Jensen--Shannon distance between each pair of regions (Figure~\ref{fig:structural_analysis})B. Higher distance means the tokenizer produces more distinct length profiles across genomic contexts.

\paragraph{Main result.} EvoLen produces higher region-to-region separation than baseline BPE for all three comparisons: promoter--enhancer increases from 0.0245 to 0.0361 ($+47\%$), promoter--exon from 0.0173 to 0.0265 ($+53\%$), and enhancer--exon from 0.0100 to 0.0141 ($+41\%$). This indicates that EvoLen assigns more distinct tokenization patterns to different functional regions. Overall, these results support \textbf{(P2) Regulatory Specificity}: EvoLen better distinguishes genomic contexts through its segmentation behavior. Full token-length distributions and pairwise divergences are reported in Appendix~\ref{sec:region_tokenization}.

\subsection{Evolutionary Conservation Alignment}

\paragraph{Why this matters.} Some genomic regions are more evolutionarily constrained than others: conserved regions tend to preserve function across species, while accelerated regions evolve more rapidly \citep{kircher2014general, gulko2015fitcons}. Because EvoLen uses evolutionary information during tokenizer construction, we expect its tokens to align more closely with these constraint patterns. This directly tests \textbf{(P3) Evolutionary Consistency}.

\paragraph{How we evaluate it.} We tokenize the same genome with baseline BPE and EvoLen, then ask how evolutionarily coherent the resulting tokens are. For each token, we average the phyloP scores of all bases it spans, and then group tokens by whether they come from conserved, neutral, or accelerated regions (Figure~\ref{fig:structural_analysis})C. If a tokenizer better respects evolutionary structure, its tokens should have higher mean phyloP in conserved regions, more negative mean phyloP in accelerated regions, and clearer separation between categories. We also examine intra-token phyloP variance to test whether any improvement reflects better alignment rather than greater heterogeneity within tokens.

\paragraph{Main result.} EvoLen produces tokens whose average conservation scores better match the evolutionary category they come from. In conserved regions, EvoLen tokens have consistently higher mean phyloP than baseline, with gains of $+1.8\%$ to $+2.3\%$ across vocabulary sizes. In neutral regions, the improvement is larger, at $+11$--$14\%$, suggesting that EvoLen better groups weakly constrained sequence patterns that baseline BPE tends to fragment. In accelerated regions, EvoLen yields slightly more negative mean phyloP values than baseline, indicating clearer separation from conserved sequence. Intra-token phyloP variance remains comparable between methods, showing that these gains come from better evolutionary alignment rather than noisier token composition. Together, these results support \textbf{(P3) Evolutionary Consistency}: EvoLen produces tokens that better respect the underlying evolutionary structure of the genome. Per-region phyloP statistics across all vocabulary sizes are reported in Appendix~\ref{sec:phylop_details}.

\subsection{Functional Sequence Enrichment}

\paragraph{Why this matters.} A tokenizer can learn frequent substrings without learning biologically meaningful ones. To test whether EvoLen captures recurrent functional sequence structure instead of generic high-frequency patterns, we ask whether its tokens show clearer enrichment in biologically distinct genomic contexts. This evaluates \textbf{(P4) Pattern Recurrence}.

\paragraph{How we evaluate it.} We first divide the genome into 12 sequence bins by crossing four genomic regions (promoter, enhancer, exon, intron) with three conservation categories (conserved, neutral, accelerated). For each bin, we tokenize all sequences assigned to that region--conservation combination and compute token frequencies. We then compare each bin against a fixed background consisting of \textbf{neutral and intronic sequences}, which serve as a broad baseline with relatively weak functional and evolutionary constraint (Figure~\ref{fig:sequence_enrichment}). We summarize this comparison using the mean $\log_2$ fold-change of token frequencies relative to that neutral intronic background. Less negative values indicate relative enrichment in a given context, whereas more negative values indicate stronger depletion. A biologically meaningful tokenizer should show clearer separation between conserved functional regions and accelerated sequence. Enrichment computation details and diagnostics are provided in Appendix~\ref{sec:enrichment_details}.

{\setlength{\intextsep}{4pt}%
\setlength{\columnsep}{8pt}%
\begin{wrapfigure}{r}{0.50\columnwidth}
    \vspace{-6mm}
    \centering
    \includegraphics[width=\linewidth, trim=0 5 0 5, clip]{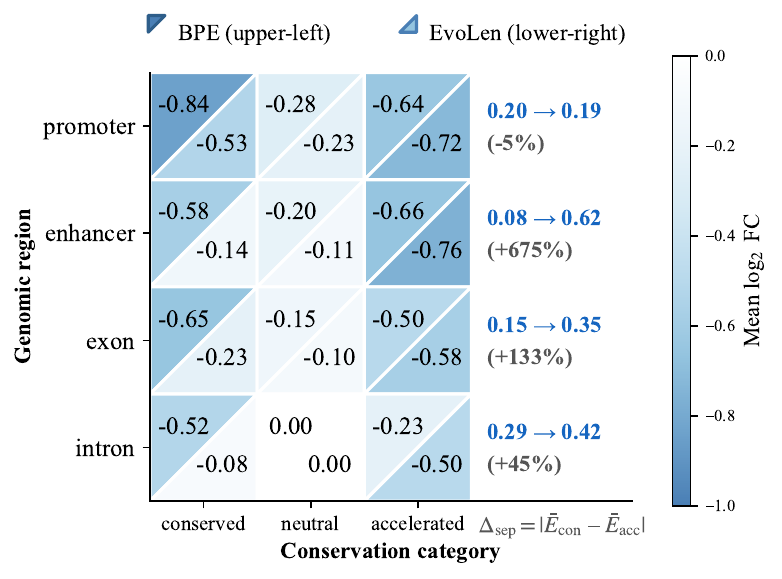}
    \vspace{-6mm}
    \caption{Token enrichment (mean $\log_2$ fold-change) relative to neutral intronic background, crossed by genomic region and conservation category(P4).}
    \label{fig:sequence_enrichment}
    \vspace{-2mm}
\end{wrapfigure}

\paragraph{Main result.} EvoLen shows stronger context-specific enrichment than baseline. In conserved regions, depletion becomes substantially weaker for promoters (from $-0.84$ to $-0.53$), enhancers (from $-0.58$ to $-0.14$), and exons (from $-0.65$ to $-0.23$), indicating that EvoLen better captures the sequence patterns characteristic of functional elements under constraint. EvoLen also increases the separation between conserved and accelerated sequence in most regions, where we define separation as $\Delta_{\mathrm{sep}} = \left| \bar{E}_{\mathrm{conserved}} - \bar{E}_{\mathrm{accelerated}} \right|$. Enhancer separation rises from $0.08$ to $0.62$, exon separation from $0.15$ to $0.35$, and intron separation from $0.29$ to $0.42$, while promoter separation remains comparable ($0.20$ vs.\ $0.19$). Overall, these results support \textbf{(P4) Pattern Recurrence}: EvoLen captures recurrent functional sequence structure more clearly than a purely frequency-driven tokenizer.

\vspace{-2mm}
\section{Downstream Evaluation}

Section~4 showed that EvoLen better preserves functional sequence units, aligns more closely with evolutionary constraint, and produces more distinctive region-specific tokenization patterns. We next ask whether these structural improvements translate into downstream predictive performance. Unless otherwise specified, all comparisons use the same model architecture, datasets, and training procedure, with models first pretrained using masked language modeling in a BERT-style setting on DNA sequences from the hg38 human reference genome and then fine-tuned and evaluated on downstream tasks.

\paragraph{Vocabulary Size Selection} Based on the token analyses in Section~4, we select \textbf{5,120} as the vocabulary size for downstream experiments, as this setting provides the best overall balance between structural fidelity and vocabulary efficiency. Per-vocabulary structural metrics are reported in Tables~\ref{tab:S1_motif_preservation}, \ref{tab:S2_phylop_combined}, \ref{tab:S4_prom_enh} and \ref{tab:S5_exon_intron_summary}.

\subsection{Controlled Tokenizer Comparison}

We compare EvoLen against standard BPE across three benchmark suites: Genome Understanding Evaluation (GUE) \citep{zhou2023dnabert2}, Genomic Benchmarks (GBM) \citep{gresova2023genomic_benchmarks}, and Nucleotide Transformer (NT) \citep{dallatorre2025nucleotide}. We also evaluate two additional regulatory settings: Multi-SCREEN, a multiclass cis-regulatory element (cCRE) classification task \citep{encode2020expanded}, and snATAC-seq, a cross-species brain cell-type cCRE classification task using human training and mouse evaluation \citep{zemke2023neocortex_regulatory_programs}. We report performance using Matthews correlation coefficient (MCC), averaged within each benchmark task group \citep{chicco2020advantages}.

\paragraph{Results.}
Table~\ref{tab:baseline_tokenizer} reports average MCC across all benchmark groups. EvoLen outperforms the baseline on 11 of 15 task groups, with the clearest gains on tasks involving regulatory structure and cross-species generalization: mouse enhancers in GenomicBench ($+9.83\%$), snATAC-seq cross-species brain cell-type prediction ($+9.47\%$), TF binding in GUE ($+5.70\%$), and mouse classification in GUE ($+4.07\%$). Consistent but smaller improvements appear on promoter-300 prediction in GUE ($+2.71\%$) and across NT histone, enhancer, and promoter prediction ($+1.88$--$2.48\%$). Across the 56 paired tasks, EvoLen improves over baseline BPE on 40 (mean $\Delta$MCC $+0.93$, median $+0.56$; bootstrap 95\% CI $[+0.34, +1.54]$ over 10,000 resamples; Wilcoxon signed-rank $p = 1.8\times10^{-3}$), indicating gains that are modest in magnitude but stable across tasks. These gains are also broadly consistent with the token-level properties identified in Section~4: better motif preservation (P1) aligns with TF-binding improvements \citep{avsec2021bpnet}, stronger region-specific tokenization patterns (P2) align with enhancer and promoter gains, and improved evolutionary alignment (P3) aligns with gains on mouse and cross-species brain cell-type prediction tasks. Although splice sites are evolutionarily conserved, the decline in performance on these tasks in both GUE and NT likely reflects their reliance on highly localized motifs and precise exon--intron boundary signals \citep{shrikumar2018modisco}, where exact local matching matters more than broader variable-length grouping. Notably, this gap narrows under compute-matched pretraining: at 200k steps the GUE splice deficit closes entirely ($+0.1\%$) and the NT splice deficit shrinks to $-0.9\%$ (Table~\ref{tab:baseline_tokenizer}), suggesting the limitation is partly a matter of convergence rather than a fixed property of the tokenizer. Invertebrate tasks also decline slightly ($-0.39\%$), and invertebrate and yeast settings likely benefit less because they fall outside the mammalian evolutionary scope used during tokenizer construction. Taken together, these results indicate that EvoLen provides meaningful improvements, with the clearest benefits on regulatory and cross-species tasks that match the functional and evolutionary biases encoded by the tokenizer.

\subsection{Tokenizer Strategy Comparison}

We further compare EvoLen with tokenization strategies used in prior DNA language models, including subword tokenization (DNABERT2-style) \citep{zhou2023dnabert2}, Nucleotide Transformer tokenization \citep{dallatorre2025nucleotide}, and GROVER tokenization \citep{sanabria2024grover}. These comparisons use a shared modeling pipeline. At 100k steps, EvoLen ranks among the top two of the four tokenization strategies on 13 of 15 task groups, despite training for half as long. Extending EvoLen and the baseline to 200k steps gives a compute-matched comparison: EvoLen improves on 12 of 15 groups, and relative to Grover-200K it reaches a 52-task mean MCC of 59.14 vs.\ 58.44 ($\Delta = +0.70$) and a 13-group mean of 56.95 vs.\ 55.07 ($\Delta = +1.88$), both excluding the four splice tasks, together with a GBM suite mean of 72.16 vs.\ 69.11 across all nine GBM tasks ($\Delta = +3.04$, paired Cohen's $d = 0.37$). EvoLen ranks in the top two among the four tokenization strategies on 94.2\% of non-splice tasks (49 of 52) and on 96.9\% of regulatory and cross-species tasks (31 of 32). Per-task results across all 56 tasks are reported in Appendix~\ref{sec:per_task_results}. Throughout, DNAbert2, NT and Grover denote the tokenization schemes introduced by those models; in every case we pretrain and fine-tune our own BERT-style model on hg38 using that tokenizer, rather than evaluating the released models themselves.

\begin{table*}[t]
\centering
\scriptsize
\setlength{\tabcolsep}{4pt}
\renewcommand{\arraystretch}{0.95}
\begin{tabular}{lc|cccccc|ccc|cccc|c|c}
\toprule
& & \multicolumn{6}{c|}{GUE} & \multicolumn{3}{c|}{GBM} & \multicolumn{4}{c|}{NT} & & \\
& Steps & EMP & Mou & P3 & PC & Spl & TF & HR & ME & Inv & His & Enh & Pro & Spl & cCRE & ATAC \\
\midrule
\multicolumn{17}{l}{\textit{Tokenizer Comparison}} \\
DNAbert2 & 200k & 46.2 & 53.6 & 76.1 & 59.1 & 73.5 & 59.0 & 68.7 & 48.6 & 69.9 & 54.5 & 46.8 & 72.9 & 69.6 & 22.0 & 8.8 \\
NT & 200k & 40.4 & 46.2 & \textbf{79.1} & 63.4 & 75.3 & 53.9 & 62.4 & 50.5 & 67.8 & 52.2 & 46.8 & 73.4 & \textbf{93.9} & 21.3 & 9.8 \\
Grover & 200k & \textbf{48.2} & \textbf{60.8} & 76.3 & 62.3 & \textbf{79.9} & \textbf{62.4} & \textbf{72.0} & 47.7 & 71.4 & 54.8 & 47.9 & 73.5 & \underline{79.6} & \textbf{23.9} & \textbf{14.5} \\
\midrule
\multicolumn{17}{l}{\textit{Baseline Comparison}} \\
Base & 100k & 46.8 & 56.5 & 76.5 & 64.2 & \underline{78.5} & 57.8 & 69.9 & 58.1 & \underline{76.4} & 53.7 & 48.7 & 74.7 & 69.3 & 21.9 & 12.2 \\
EvoLen & 100k & 46.8 & 58.8 & \underline{78.5} & 63.6 & 75.2 & 61.0 & 70.3 & \underline{63.8} & 76.1 & \underline{55.0} & 49.6 & \textbf{76.5} & 67.8 & 22.1 & 13.3 \\
$\Delta$\% & & \textbf{+0.1} & \textbf{+4.1} & \textbf{+2.7} & -1.0 & -4.2 & \textbf{+5.7} & \textbf{+0.5} & \textbf{+9.8} & -0.4 & \textbf{+2.3} & \textbf{+1.9} & \textbf{+2.5} & -2.2 & \textbf{+1.0} & \textbf{+9.5} \\
\midrule
\multicolumn{17}{l}{\textit{Compute-Matched Comparison}} \\
Base & 200k & 46.9 & 54.5 & 77.6 & \textbf{64.8} & 75.5 & 58.6 & 69.9 & 53.8 & 76.2 & 53.5 & \underline{49.7} & 74.9 & 69.0 & \underline{22.7} & 10.3 \\
EvoLen & 200k & \underline{47.9} & \underline{58.9} & 78.1 & \underline{64.4} & 75.6 & \underline{61.2} & \underline{70.8} & \textbf{64.2} & \textbf{77.0} & \textbf{55.4} & \textbf{50.0} & \underline{76.3} & 68.4 & 22.5 & \underline{13.7} \\
$\Delta$\% & & \textbf{+2.1} & \textbf{+8.1} & \textbf{+0.6} & -0.6 & \textbf{+0.1} & \textbf{+4.4} & \textbf{+1.3} & \textbf{+19.3} & \textbf{+1.0} & \textbf{+3.6} & \textbf{+0.6} & \textbf{+1.9} & -0.9 & -0.9 & \textbf{+33.0} \\
\bottomrule
\end{tabular}
\caption{
Average MCC (\%) across downstream benchmarks. \textbf{Bold} = best; \underline{underline} = second-best, computed across all seven model rows (including both step counts of Base and EvoLen).
 $\Delta$\% = relative improvement of EvoLen over baseline BPE. Steps = number of pretraining steps. Fine-tuning hyperparameters are selected per task. Column abbreviations --- GUE: EMP = EMP (Yeast), Mou = Mouse, P3 = Promoter-300, PC = Promoter-Core, Spl = Splice, TF = TF binding; GBM: HR = Human regulatory, ME = Mouse enhancers, Inv = Invertebrates; NT: His = Histone marks, Enh = Enhancers, Pro = Promoters, Spl = Splice; cCRE = Multi-SCREEN (Human cCRE); ATAC = snATAC-seq (Human/Mouse brain).
}
\label{tab:baseline_tokenizer}
\end{table*}

\subsection{Ablation Study}
We ablate the three main components of EvoLen at vocabulary size 5{,}120: three-way evolutionary partitioning, conservation-prioritized merging, and length-aware scoring. Functional and evolutionary token analysis results are reported in Appendix Table~\ref{tab:ablation_structural}, and downstream analysis results are provided in Appendix Table~\ref{tab:ablation_downstream}. Among the tokenizer-construction components, conservation-prioritized merging has the largest effect on structural quality, while evolutionary partitioning provides a smaller but consistent benefit. In downstream evaluation, full EvoLen performs best on nearly all task groups, with snATAC-seq essentially tied with the \textbf{No Partition} variant (13.3 vs.\ 13.4 MCC, a negligible difference relative to the overall difficulty of cross-species generalization at this scale). Overall, these results suggest that most of EvoLen's gains come from selecting biologically informative tokens during vocabulary construction, while length-aware decoding provides a refinement. 

\vspace{-2mm}
\section{Related Work}

\paragraph{DNA language models and tokenization.}
Recent genomic foundation models have shown that large-scale pretraining can learn useful sequence representations from DNA \citep{ji2021dnabert, zhou2023dnabert2, nguyen2024evo}. Existing tokenization strategies largely fall into three categories: fixed-length $k$-mers, adaptive subword tokenization, and single-nucleotide resolution. Fixed-length approaches like DNABERT \citep{ji2021dnabert} impose a uniform scale that can fragment regulatory elements, while adaptive subword methods such as DNABERT2 and GROVER \citep{zhou2023dnabert2, sanabria2024grover} remain driven primarily by frequency rather than biological function. At the other extreme, nucleotide-level models such as Evo and HyenaDNA \citep{nguyen2024evo, hyenadna2023} avoid fragmentation but sacrifice the efficiency of learned variable-length subwords \citep{clark2022canine}. EvoLen addresses this gap.

\paragraph{Biological priors and evolutionary information.}
Biological sequence function is organized across multiple scales, from short TF-binding motifs to broader regulatory annotations \citep{jaspar2024, encode2020expanded}. Evolutionary conservation has long served as a proxy for functional importance, and scores such as phyloP provide nucleotide-level estimates of selective constraint across species \citep{siepel2010evolutionary, ucsc_phylop}. Recent large-scale sequence models can absorb such signals implicitly through broad phylogenetic training corpora \citep{nguyen2024evo, brixi2026evo2}. In contrast, EvoLen incorporates conservation explicitly into subword vocabulary construction and decoding. To our knowledge, prior DNA tokenizers have not incorporated evolutionary conservation directly into subword vocabulary design. EvoLen is complementary to both scaling-oriented DNA language models and existing benchmark efforts \citep{genome_benchmark2025, gresova2023genomic_benchmarks}.

\vspace{-2mm}
\section{Conclusion}

Our results demonstrate that tokenization is not a neutral preprocessing step for DNALMs: it introduces a biologically informed inductive bias that shapes how regulatory sequence patterns are represented and learned. By incorporating evolutionary constraint directly into vocabulary construction and decoding, EvoLen provides this bias without requiring changes to model architecture or training objectives. Across token analyses, EvoLen improves motif preservation, evolutionary alignment, and regulatory element-level differentiation, and these gains translate most clearly to downstream improvements on tasks involving regulatory modeling and cross-species generalization. More broadly, for DNALM, structured tokenization offers a lightweight path to better representations, one that complements, rather than competes with, gains from increased model scale.

\paragraph{Limitations and Future Directions.}
EvoLen's benefits are strongest when downstream tasks align with the evolutionary scope of the tokenizer, and the gains are less consistent for tasks dominated by exact short-boundary recognition or for species outside the conservation signal used during tokenizer construction. We assess robustness at the task level rather than across pretraining seeds; multi-seed pretraining is left to future work. A promising next step is to extend EvoLen beyond evolutionary conservation by incorporating additional biological priors into token construction, such as motif coverage, GC context, or other sequence-level features. Another important direction is to test whether biologically informed tokenization yields larger benefits in downstream applications such as sequence design and individual genome annotation \citep{zhou2015deepsea}. Variant effect prediction is a particularly direct test, since it probes whether a tokenizer's units align with functional sequence elements; it is not a standard benchmark for BERT-style DNALMs --- the GUE, GBM and NT suites are classification-oriented --- and requires methodology distinct from sequence classification, so we leave it to future work.

\section*{Disclosure of LLM Usage}

In accordance with Policy 1 regarding the use of Large Language Models (LLMs) in research and reviewing, we disclose the following:

\begin{itemize}
    \item \textbf{Research Content Generation:} No LLMs were used to originate research ideas, write original text, or generate data.
\end{itemize}

\textbf{Accountability Statement:} The authors acknowledge that while LLMs may have been used as assistive tools, the human authors remain fully accountable for the final content. We have independently verified all citations, data points, and technical claims. We accept full responsibility for the accuracy, integrity, and originality of this work; any inaccuracies or hallucinations resulting from LLM usage are the sole responsibility of the authors.

\bibliography{colm2026_conference}

@article{zhou2023dnabert2,
  title={DNABERT-2: Efficient Foundation Model and Benchmark for Multi-Species Genome},
  author={Zhou, Zhihan and Ji, Yanrong and Li, Weijian and Dutta, Pratik and Davuluri, Ramana and Liu, Han},
  journal={arXiv preprint arXiv:2306.15006},
  year={2023},
  doi={10.48550/arXiv.2306.15006}
}

@article{ji2021dnabert,
  title={DNABERT: Pre-trained Bidirectional Encoder Representations from Transformers Model for DNA-Language in Genome},
  author={Ji, Yanrong and Zhou, Zhihan and Liu, Han and Davuluri, Ramana},
  journal={Bioinformatics},
  volume={37},
  number={15},
  pages={2112--2120},
  year={2021},
  publisher={Oxford University Press},
  doi={10.1093/bioinformatics/btab083}
}

@article{dallatorre2025nucleotide,
  title={Nucleotide Transformer: Building and Evaluating Robust Foundation Models for Human Genomics},
  author={Dalla-Torre, Hugo and Gonzalez, Liam and Mendoza-Revilla, Javier and others},
  journal={Nature Methods},
  volume={22},
  number={2},
  pages={287--297},
  year={2025},
  doi={10.1038/s41592-024-02523-z}
}

@article{hyenadna2023,
  title={HyenaDNA: Long-Range Genomic Sequence Modeling at Single Nucleotide Resolution},
  author={Nguyen, Eric and Poli, Michael and Faizi, Marjan and Thomas, Armin and others},
  journal={arXiv preprint arXiv:2306.15794},
  year={2023}
}

@article{encode2020expanded,
  title={Expanded encyclopaedias of DNA elements in the human and mouse genomes},
  author={{ENCODE Project Consortium} and Moore, Jill E. and Purcaro, Michael J. and Pratt, Henry E. and Epstein, Charles B. and Shoresh, Noam and Adrian, Jessika and others},
  journal={Nature},
  volume={583},
  number={7818},
  pages={699--710},
  year={2020},
  doi={10.1038/s41586-020-2493-4},
  url={https://doi.org/10.1038/s41586-020-2493-4}
}

@article{genome_benchmark2025,
  title={Benchmarking DNA foundation models for genomic and genetic tasks},
  author={Feng, Haonan and Wu, Lang and Zhao, Bingxin and others},
  journal={Nature Communications},
  volume={16},
  number={1},
  pages={10780},
  year={2025},
  doi={10.1038/s41467-025-65823-8}
}

@misc{ucsc_genome_browser,
  title={UCSC Genome Browser Downloads},
  author={{UCSC Genome Browser}},
  year={2024},
  url={https://hgdownload.gi.ucsc.edu/downloads.html}
}

@misc{ucsc_phylop,
  title={PhyloP 20-way Conservation Scores (hg38)},
  author={{UCSC Genome Browser}},
  year={2024},
  url={https://hgdownload.cse.ucsc.edu/goldenpath/hg38/phyloP20way/}
}

@article{seizl2011ga_element,
  title={A Conserved GA Element in TATA-Less RNA Polymerase II Promoters},
  author={Seizl, Martin and Hartmann, Holger and Hoeg, Friederike and Kurth, Fabian and Martin, Dietmar E. and S{\"o}ding, Johannes and Cramer, Patrick},
  journal={PLoS ONE},
  volume={6},
  number={11},
  pages={e27595},
  year={2011},
  doi={10.1371/journal.pone.0027595},
  pmcid={PMC3217976},
  pmid={22110682},
  url={https://pmc.ncbi.nlm.nih.gov/articles/PMC3217976/}
}

@article{sanabria2024grover,
  title={DNA language model GROVER learns sequence context in the human genome},
  author={Sanabria, Melissa and Hirsch, Jonas and Joubert, Pierre M. and others},
  journal={Nature Machine Intelligence},
  volume={6},
  pages={911--923},
  year={2024},
  doi={10.1038/s42256-024-00872-0},
  url={https://doi.org/10.1038/s42256-024-00872-0}
}

@article{gresova2023genomic_benchmarks,
  title={Genomic benchmarks: a collection of datasets for genomic sequence classification},
  author={Gre{\v{s}}ov{\'a}, Katar{\'\i}na and Martinek, Vlastimil and {\v{C}}ech{\'a}k, David and {\v{S}}ime{\v{c}}ek, Petr and Alexiou, Panagiotis},
  journal={BMC Genomic Data},
  volume={24},
  pages={25},
  year={2023},
  doi={10.1186/s12863-023-01123-8},
  pmcid={PMC10150520},
  pmid={37127596},
  url={https://pmc.ncbi.nlm.nih.gov/articles/PMC10150520/}
}

@article{zemke2023neocortex_regulatory_programs,
  title={Conserved and divergent gene regulatory programs of the mammalian neocortex},
  author={Zemke, Nathan R. and Armand, Ethan J. and Wang, Wenliang and others},
  journal={Nature},
  volume={624},
  number={7991},
  pages={390--402},
  year={2023},
  doi={10.1038/s41586-023-06819-6},
  url={https://doi.org/10.1038/s41586-023-06819-6}
}

@article{siepel2010evolutionary,
  title={Detection of nonneutral substitution rates on mammalian phylogenies},
  author={Pollard, Katherine S. and  Hubisz, Melissa J. and Rosenbloom, Kate R. and Siepel, Adam},
  journal={Genome Research},
  volume={20},
  number={1},
  pages={110--121},
  year={2010},
  publisher={Cold Spring Harbor Lab},
  doi={10.1101/gr.097857.109}
}

@article{zhou2015deepsea,
  title={Predicting effects of noncoding variants with deep learning--based sequence model},
  author={Zhou, Jian and Troyanskaya, Olga G},
  journal={Nature Methods},
  volume={12},
  number={10},
  pages={931--934},
  year={2015},
  doi={10.1038/nmeth.3547}
}

@article{jaspar2024,
  title={{JASPAR 2024}: 20th anniversary of the open-access database of transcription factor binding profiles},
  author={Rauluseviciute, Ieva and others},
  journal={Nucleic Acids Research},
  volume={52},
  number={D1},
  pages={D174--D182},
  year={2024},
  doi={10.1093/nar/gkad1059}
}

@article{brixi2026evo2,
  title={Genome modeling and design across all domains of life with {E}vo 2},
  author={Brixi, Garyk and Durrant, Matthew G. and Ku, Jerome and Naghipourfar, Mohsen and Poli, Michael and Hie, Brian L. and others},
  journal={Nature},
  year={2026},
  doi={10.1038/s41586-026-10176-5},
  url={https://doi.org/10.1038/s41586-026-10176-5}
}

@article{nguyen2024evo,
  title={Sequence modeling and design from molecular to genome scale with {E}vo},
  author={Nguyen, Eric and Poli, Michael and Durrant, Matthew G. and Thomas, Armin W. and others},
  journal={Science},
  volume={386},
  number={6723},
  pages={eado9336},
  year={2024},
  publisher={American Association for the Advancement of Science},
  doi={10.1126/science.ado9336}
}

@article{siepel2005phastcons,
  title={Evolutionarily conserved elements in vertebrate, insect, worm, and yeast genomes},
  author={Siepel, Adam and Bejerano, Gill and Pedersen, Jakob S and Hinrichs, Aris S and Hou, Minmei and Rosenbloom, Stephanie T and Clawson, Hiram and Spieth, John and Hillier, LaDeana W and Richards, Stephen and others},
  journal={Genome Research},
  volume={15},
  number={8},
  pages={1034--1050},
  year={2005},
  publisher={Cold Spring Harbor Lab},
  doi={10.1101/gr.3715005}
}

@article{davydov2010gerp,
  title={Identifying a high fraction of the human genome to be under selective constraint using GERP++},
  author={Davydov, Eugene V and Goode, David L and Sirota, Marina and Cooper, Gregory M and Sidow, Arend and Batzoglou, Serafim},
  journal={PLoS Computational Biology},
  volume={6},
  number={12},
  pages={e1001025},
  year={2010},
  publisher={Public Library of Science},
  doi={10.1371/journal.pcbi.1001025}
}

@inproceedings{kudo2018subword,
  title={Subword Regularization: Improving Neural Network Translation Models with Multiple Subword Candidates},
  author={Kudo, Taku},
  booktitle={Proceedings of the 56th Annual Meeting of the Association for Computational Linguistics (Volume 1: Long Papers)},
  pages={66--75},
  year={2018},
  doi={10.18653/v1/P18-1007}
}

@inproceedings{kudo2018sentencepiece,
  title={{S}entence{P}iece: A simple and language independent subword tokenizer and detokenizer for Neural Text Processing},
  author={Kudo, Taku and Richardson, John},
  booktitle={Proceedings of the 2018 Conference on Empirical Methods in Natural Language Processing: System Demonstrations},
  pages={66--71},
  year={2018}
}

@article{provilkov2019bpe,
  title={BPE-Dropout: Simple and Effective Subword Regularization},
  author={Provilkov, Ivan and Emelianenko, Dmitrii and Voita, Elena},
  journal={arXiv preprint arXiv:1910.13267},
  year={2019}
}

@article{clark2022canine,
  title={CANINE: Pre-training an Efficient Tokenization-free Encoder for Language Representation},
  author={Clark, Jonathan H and Garrette, Dan and Turc, Iulia and Wieting, John},
  journal={Transactions of the Association for Computational Linguistics},
  volume={10},
  pages={73--91},
  year={2022}
}

@inproceedings{schmidt2024tokenization,
  title     = {Tokenization Is More Than Compression},
  author    = {Schmidt, Craig W. and Reddy, Varshini and Zhang, Haoran and
               Alameddine, Alec and Uzan, Omri and Pinter, Yuval and
               Tanner, Chris},
  booktitle = {Proceedings of the 2024 Conference on Empirical Methods in Natural Language Processing},
  month     = nov,
  year      = {2024},
  address   = {Miami, Florida, USA},
  publisher = {Association for Computational Linguistics},
  pages     = {678--702},
  doi       = {10.18653/v1/2024.emnlp-main.40},
  url       = {https://aclanthology.org/2024.emnlp-main.40/}
}

@inproceedings{sennrich2016neural,
  title     = {Neural Machine Translation of Rare Words with Subword Units},
  author    = {Sennrich, Rico and Haddow, Barry and Birch, Alexandra},
  booktitle = {Proceedings of the 54th Annual Meeting of the Association for
               Computational Linguistics (Volume 1: Long Papers)},
  pages     = {1715--1725},
  year      = {2016},
  address   = {Berlin, Germany},
  publisher = {Association for Computational Linguistics},
  doi       = {10.18653/v1/P16-1162},
  url       = {https://aclanthology.org/P16-1162/}
}

@article{kircher2014general,
  title={A general framework for estimating the relative pathogenicity of human genetic variants},
  author={Kircher, Martin and Witten, Daniela M and Jain, Preti and O'Roak, Brian J and Cooper, Gregory M and Shendure, Jay},
  journal={Nature Genetics},
  volume={46},
  number={3},
  pages={310--315},
  year={2014},
  doi={10.1038/ng.2892}
}

@article{gulko2015fitcons,
  title={A method for calculating probabilities of fitness consequences for point mutations across the human genome},
  author={Gulko, Brad and Hubisz, Melissa J and Gronau, Ilan and Siepel, Adam},
  journal={Nature Genetics},
  volume={47},
  number={3},
  pages={276--283},
  year={2015}
}

@article{alipanahi2015predicting,
  title={Predicting the sequence specificities of DNA- and RNA-binding proteins by deep learning},
  author={Alipanahi, Babak and Delong, Andrew and Weirauch, Matthew T and Frey, Brendan J},
  journal={Nature Biotechnology},
  volume={33},
  number={8},
  pages={831--838},
  year={2015}
}

@article{kelley2018sequential,
  title={Sequential regulatory activity prediction across chromosomes with convolutional neural networks},
  author={Kelley, David R and Reshef, Yakir A. and Bileschi, Maxwell and Belanger, David and others},
  journal={Genome Research},
  volume={28},
  number={5},
  pages={739--750},
  year={2018}
}

@article{bailey2009meme,
  title={MEME SUITE: tools for motif discovery and searching},
  author={Bailey, Timothy L and Boden, Mikael and Buske, Fabian A and others},
  journal={Nucleic Acids Research},
  volume={37},
  number={suppl\_2},
  pages={W202--W208},
  year={2009}
}

@article{grant2011fimo,
  title={FIMO: scanning for occurrences of a given motif},
  author={Grant, Charles E and Bailey, Timothy L and Noble, William S},
  journal={Bioinformatics},
  volume={27},
  number={7},
  pages={1017--1018},
  year={2011}
}

@article{dnamotiftokenizer2025,
  title={DNAMotifTokenizer: Towards Biologically Informed Tokenization of Genomic Sequences},
  author={Zhou, Xiaoxiao and Wang, Zihan and Shang, Jingbo and Li, Yang E.},
  journal={arXiv preprint arXiv:2512.17126},
  year={2025}
}

@article{avsec2021bpnet,
  title={Base-resolution models of transcription-factor binding reveal soft motif syntax},
  author={Avsec, {\v{Z}}iga and Weilert, Melanie and Shrikumar, Avanti and Krueger, Sabrina and Alexandari, Amr and Dalal, Khyati and Fropf, Robin and McAnany, Charles and Gagneur, Julien and Kundaje, Anshul and others},
  journal={Nature Genetics},
  volume={53},
  number={3},
  pages={354--366},
  year={2021},
  doi={10.1038/s41588-021-00782-6}
}

@article{shrikumar2018modisco,
  title={Technical Note on Transcription Factor Motif Discovery from Importance Scores (TF-MoDISco)},
  author={Shrikumar, Avanti and others},
  journal={arXiv preprint arXiv:1811.00416},
  year={2018}
}

@article{chicco2020advantages,
  title={The advantages of the Matthews correlation coefficient (MCC) over F1 score and accuracy in binary classification evaluation},
  author={Chicco, Davide and Jurman, Giuseppe},
  journal={BMC Genomics},
  volume={21},
  pages={6},
  year={2020},
  doi={10.1186/s12864-019-6413-7}
}
\bibliographystyle{colm2026_conference}

\newpage
\vspace{-2mm}
\appendix
\section{Appendix}
\subsection{Additional Motif Preservation Analysis}

\begin{figure}[htbp]
\centering
\includegraphics[width=\linewidth, trim=0 5 0 5, clip]{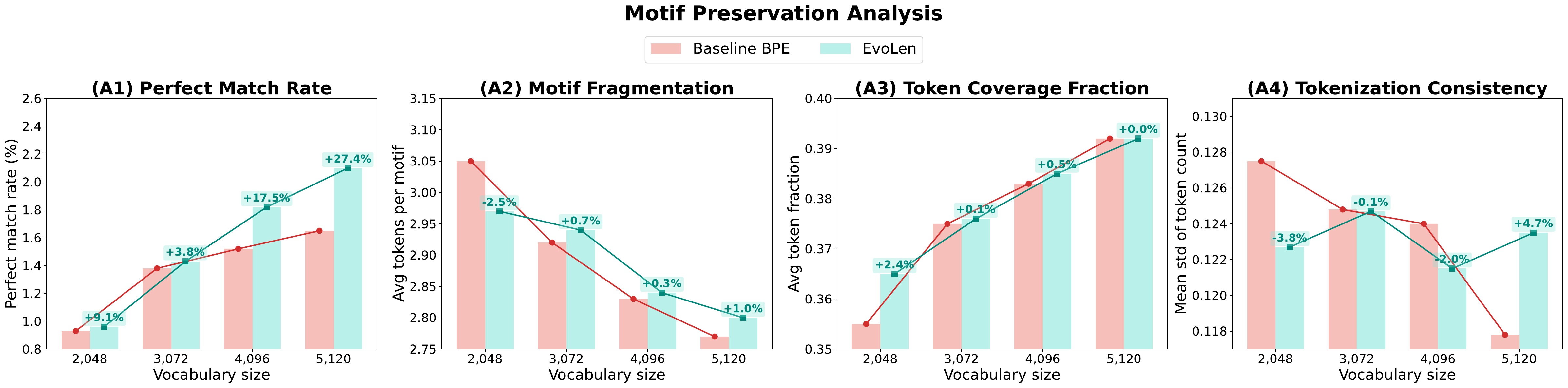}
\vspace{-4mm}
\caption{Motif preservation diagnostics across vocabulary sizes: perfect match rate, fragmentation, token coverage fraction, and tokenization consistency. EvoLen improves exact motif preservation while maintaining comparable performance on other metrics.}
\label{fig:motif_preservation_appendix}
\end{figure}

Since \textbf{Motif fragmentation} (average tokens per motif) is heavily influenced by motif length---longer motifs naturally require more tokens regardless of the tokenizer---we report two complementary metrics. \textbf{Token Coverage Fraction} normalizes by motif length and measures the average fraction of the motif covered by each token (higher = fewer, longer-spanning tokens). \textbf{Tokenization Consistency} measures the standard deviation of token counts across wildcard-expanded variants of the same motif family (lower = more stable segmentation). Formally, for a tokenizer $t$ and motif family $m$, let $V(m)$ denote the set of all expanded variants of $m$, and let $n_{m,v}$ denote the number of tokens produced when encoding variant $v \in V(m)$. We define:
\[
\mathrm{Consistency}(t)=\frac{1}{|M|}\sum_{m\in M}\mathrm{std}\bigl(\{n_{m,v}\}_{v\in V(m)}\bigr),
\]
where $M$ is the set of motif families. These metrics show that EvoLen's improvement is concentrated in exact motif preservation rather than bulk fragmentation reduction (Table~\ref{tab:S1_motif_preservation}, Figure~\ref{fig:motif_preservation_appendix}). In Table~\ref{tab:S1_motif_preservation}, \textit{AvgTok/Motif} is the average number of tokens a motif is split into; \textit{PerfectMatch\%} is the fraction of motifs encoded as a single token; \textit{ExactVocab\%} is the fraction of motifs that appear verbatim in the vocabulary; and \textit{AvgTokenFrac} is the average fraction of the motif covered by each token (higher values indicate longer-spanning tokens).

\begin{figure}[htbp]
\centering
\includegraphics[width=\linewidth, trim=0 5 0 5, clip]{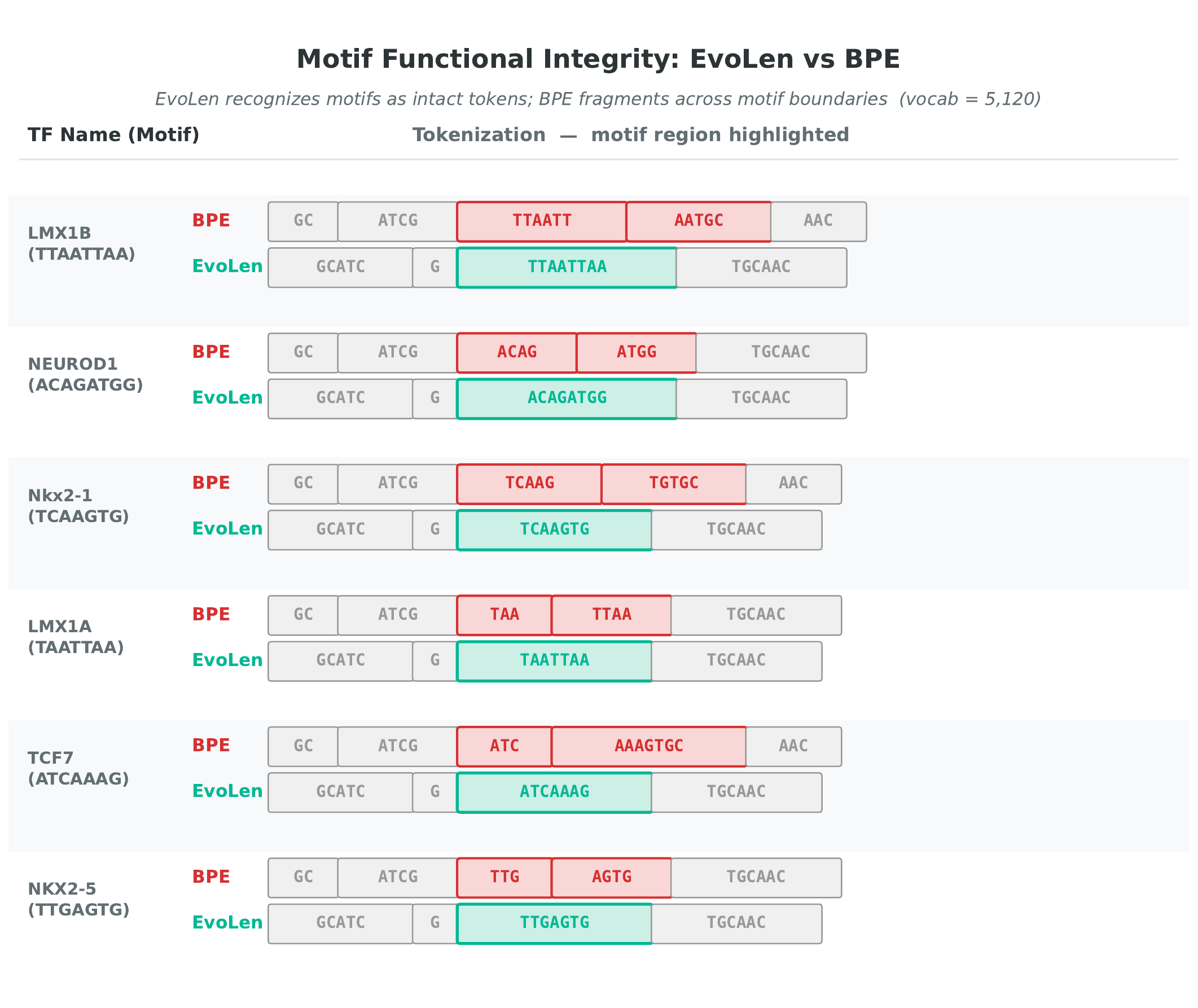}
\vspace{-4mm}
\caption{Motif fragmentation examples at vocabulary size 5,120. EvoLen preserves motifs as intact tokens, whereas BPE splits them across token boundaries.}
\label{fig:motif_fragmentation_appendix}
\end{figure}

\begin{table}[htbp]
\centering
\small
\caption{Motif preservation summary across tokenizer variants and vocabulary sizes.}\label{tab:S1_motif_preservation}
\begin{tabular}{@{}lcccc@{}}
\toprule
Tokenizer & AvgTok/Motif & PerfectMatch\% & ExactVocab\% & AvgTokenFrac \\
\midrule
Baseline BPE 5120 & 2.776 & 1.659 & 1.659 & 0.389 \\
EvoLen 5120 & 2.804 & 2.114 & 2.114 & 0.389 \\
Baseline BPE 4096 & 2.842 & 1.525 & 1.525 & 0.380 \\
EvoLen 4096 & 2.849 & 1.793 & 1.793 & 0.382 \\
Baseline BPE 3072 & 2.914 & 1.391 & 1.391 & 0.371 \\
EvoLen 3072 & 2.934 & 1.445 & 1.445 & 0.371 \\
Baseline BPE 2048 & 3.049 & 0.883 & 0.883 & 0.355 \\
EvoLen 2048 & 2.974 & 0.963 & 0.963 & 0.363 \\
Grover            & 3.404 & 0.134 & 0.134 & 0.322 \\
DNAbert2          & 2.786 & 2.435 & 2.435 & 0.393 \\
\bottomrule
\end{tabular}
\end{table}

\subsection{Evolutionary Conservation Details}\label{sec:phylop_details}

Table~\ref{tab:S2_phylop_combined} reports per-region phyloP statistics for both EvoLen and baseline BPE across vocabulary sizes. \textit{MeanPhyloP} is the mean per-token phyloP score averaged across all tokens in a given region; \textit{\%$>$0} denotes the percentage of tokens whose mean phyloP is positive, indicating net conservation; \textit{MeanVar} is the mean intra-token phyloP variance, measuring how homogeneous each token's constituent bases are in conservation level. EvoLen consistently achieves higher mean phyloP in conserved regions, larger improvements in neutral regions, and slightly more negative values in accelerated regions, indicating better alignment with evolutionary constraint.

\begin{table}[htbp]
\centering
\small
\caption{PhyloP summary by conservation category and vocabulary size, including other tokenization strategies at their native vocabulary sizes.}\label{tab:S2_phylop_combined}
\setlength{\tabcolsep}{3.5pt}
\begin{tabular}{@{}llccccc@{}}
\toprule
Method & Vocab & Region & MeanPhyloP & \%$>$0 & MeanVar & Tokens \\
\midrule
EvoLen & 2048 & conserved & 0.558 & 100.0 & 0.283 & 2,042 \\
EvoLen & 2048 & neutral & 0.080 & 87.1 & 0.348 & 2,042 \\
EvoLen & 2048 & accelerated & $-$0.273 & 0.3 & 0.635 & 2,042 \\
Baseline & 2048 & conserved & 0.549 & 100.0 & 0.283 & 2,035 \\
Baseline & 2048 & neutral & 0.072 & 84.7 & 0.347 & 2,042 \\
Baseline & 2048 & accelerated & $-$0.271 & 0.9 & 0.633 & 2,042 \\
\midrule
EvoLen & 3072 & conserved & 0.559 & 100.0 & 0.287 & 3,062 \\
EvoLen & 3072 & neutral & 0.081 & 87.7 & 0.352 & 3,066 \\
EvoLen & 3072 & accelerated & $-$0.274 & 0.3 & 0.644 & 3,066 \\
Baseline & 3072 & conserved & 0.546 & 100.0 & 0.286 & 3,041 \\
Baseline & 3072 & neutral & 0.071 & 84.9 & 0.352 & 3,066 \\
Baseline & 3072 & accelerated & $-$0.272 & 0.8 & 0.642 & 3,065 \\
\midrule
EvoLen & 4096 & conserved & 0.557 & 100.0 & 0.290 & 4,083 \\
EvoLen & 4096 & neutral & 0.080 & 87.3 & 0.356 & 4,090 \\
EvoLen & 4096 & accelerated & $-$0.275 & 0.2 & 0.651 & 4,090 \\
Baseline & 4096 & conserved & 0.545 & 100.0 & 0.289 & 4,045 \\
Baseline & 4096 & neutral & 0.070 & 84.6 & 0.355 & 4,090 \\
Baseline & 4096 & accelerated & $-$0.274 & 0.7 & 0.649 & 4,086 \\
\midrule
EvoLen & 5120 & conserved & 0.558 & 100.0 & 0.291 & 5,107 \\
EvoLen & 5120 & neutral & 0.081 & 87.2 & 0.358 & 5,114 \\
EvoLen & 5120 & accelerated & $-$0.273 & 0.2 & 0.654 & 5,114 \\
Baseline & 5120 & conserved & 0.545 & 100.0 & 0.290 & 5,042 \\
Baseline & 5120 & neutral & 0.070 & 84.7 & 0.356 & 5,109 \\
Baseline & 5120 & accelerated & $-$0.272 & 0.6 & 0.651 & 5,106 \\
\midrule
Grover & 605 & conserved & 0.549 & 100.0 & 0.271 & 604 \\
Grover & 605 & neutral & 0.071 & 84.8 & 0.329 & 604 \\
Grover & 605 & accelerated & $-$0.274 & 1.3 & 0.601 & 604 \\
DNAbert2 & 4096 & conserved & 0.548 & 100.0 & 0.294 & 4,088 \\
DNAbert2 & 4096 & neutral & 0.068 & 82.1 & 0.363 & 4,091 \\
DNAbert2 & 4096 & accelerated & $-$0.282 & 0.6 & 0.656 & 4,091 \\
\bottomrule
\end{tabular}
\end{table}

\subsection{Region-Specific Tokenization Analysis}\label{sec:region_tokenization}

\begin{table}[htbp]
\centering
\small
\caption{Token length-bin composition (\%) across four genomic regions at vocabulary sizes 3,072 and 5,120.}\label{tab:S9_four_region_dist}
\setlength{\tabcolsep}{3.5pt}
\begin{tabular}{@{}llccccc@{}}
\toprule
Method & Vocab & Region & Pct1--2 & Pct3--5 & Pct6--8 & Pct9+ \\
\midrule
Baseline & 3072 & promoter & 4.48 & 63.72 & 30.91 & 0.90 \\
Baseline & 3072 & enhancer & 4.12 & 62.00 & 33.23 & 0.66 \\
Baseline & 3072 & exon & 4.01 & 62.25 & 32.91 & 0.84 \\
Baseline & 3072 & intron & 2.15 & 61.01 & 35.32 & 1.52 \\
EvoLen & 3072 & promoter & 26.59 & 23.71 & 49.03 & 0.67 \\
EvoLen & 3072 & enhancer & 27.79 & 18.85 & 52.81 & 0.55 \\
EvoLen & 3072 & exon & 27.22 & 20.09 & 52.03 & 0.66 \\
EvoLen & 3072 & intron & 26.94 & 17.20 & 54.73 & 1.14 \\
\midrule
Baseline & 5120 & promoter & 3.27 & 58.09 & 37.35 & 1.29 \\
Baseline & 5120 & enhancer & 3.19 & 55.98 & 39.83 & 1.01 \\
Baseline & 5120 & exon & 3.04 & 56.31 & 39.40 & 1.25 \\
Baseline & 5120 & intron & 1.40 & 54.35 & 41.96 & 2.29 \\
EvoLen & 5120 & promoter & 26.19 & 17.26 & 55.20 & 1.35 \\
EvoLen & 5120 & enhancer & 27.27 & 14.18 & 57.34 & 1.21 \\
EvoLen & 5120 & exon & 26.86 & 14.90 & 56.80 & 1.44 \\
EvoLen & 5120 & intron & 26.35 & 12.91 & 58.38 & 2.36 \\
\bottomrule
\end{tabular}
\end{table}

Tables~\ref{tab:S9_four_region_dist} and~\ref{tab:S10_js} report token-length distributions across four genomic regions and pairwise Jensen--Shannon divergences at vocabulary sizes 3,072 and 5,120. Note that the Jensen--Shannon distance reported in the main text (Figure~\ref{fig:structural_analysis}B) is the square root of the Jensen--Shannon divergence; the appendix tables report the divergence directly. \textit{Pct1--2}, \textit{Pct3--5}, \textit{Pct6--8}, and \textit{Pct9+} denote the percentage of tokens whose length falls in the corresponding base-pair range. EvoLen produces more distinct length profiles across regions, with a larger fraction of tokens in the 6--8~bp range and greater inter-region separation.

\begin{table}[htbp]
\centering
\small
\caption{Pairwise Jensen--Shannon divergence between region token-length distributions at vocabulary size 5,120.}\label{tab:S10_js}
\begin{tabular}{@{}llcc@{}}
\toprule
RegionA & RegionB & Baseline & EvoLen \\
\midrule
promoter & enhancer & 0.001 & 0.001 \\
promoter & exon & 0.000 & 0.001 \\
promoter & intron & 0.005 & 0.004 \\
enhancer & exon & 0.000 & 0.000 \\
enhancer & intron & 0.005 & 0.002 \\
exon & intron & 0.004 & 0.001 \\
\bottomrule
\end{tabular}
\end{table}

\subsection{Functional Sequence Enrichment Details}\label{sec:enrichment_details}

To quantify whether a tokenizer captures region- and conservation-specific sequence patterns, we compute the mean $\log_2$ fold-change of per-token frequencies relative to a neutral intronic background. We construct 12 genomic bins by intersecting four region annotations (promoter, enhancer, exon, intron) with three conservation categories (conserved, neutral, accelerated). For a given bin $b$ and the reference background bin $\mathrm{bg}$ (intron $\times$ neutral), the smoothed frequency of token $t$ is:
\[
f_t^{(b)} = \frac{c_t^{(b)} + \alpha}{N^{(b)} + \alpha \cdot |V|}
\]
where $c_t^{(b)}$ is the raw count, $N^{(b)} = \sum_t c_t^{(b)}$, $|V|$ is the vocabulary size, and $\alpha = 0.5$. The per-token log fold-change and enrichment score are:
\[
\log_2\mathrm{FC}(t, b) = \log_2 \frac{f_t^{(b)}}{f_t^{(\mathrm{bg})}}, \qquad \bar{E}(b) = \frac{1}{|V|} \sum_{t \in V} \log_2\mathrm{FC}(t, b)
\]
Negative values indicate tokens that are less frequent than in the intronic neutral background; by definition $\bar{E}(\mathrm{intron} \times \mathrm{neutral}) = 0$. Tables~\ref{tab:S4_prom_enh} and~\ref{tab:S5_exon_intron_summary} report enrichment diagnostics for promoter--enhancer and exon--intron comparisons.

\begin{table}[H]
\centering
\small
\caption{Promoter versus enhancer token enrichment diagnostics.}\label{tab:S4_prom_enh}
\setlength{\tabcolsep}{3pt}
\begin{tabular}{@{}llcccccc@{}}
\toprule
Model & Vocab & SigTokens & SigProm & SigEnh & PromShare\% & EnhMeanLen & EnhMeanGC \\
\midrule
Baseline & 2048 & 1,736 & 456 & 1,280 & 26.27 & 6.21 & 0.337 \\
EvoLen & 2048 & 1,852 & 380 & 1,472 & 20.52 & 6.31 & 0.332 \\
Baseline & 3072 & 2,449 & 598 & 1,851 & 24.42 & 6.51 & 0.345 \\
EvoLen & 3072 & 2,746 & 517 & 2,229 & 18.83 & 6.65 & 0.336 \\
Baseline & 4096 & 3,062 & 755 & 2,307 & 24.66 & 6.69 & 0.347 \\
EvoLen & 4096 & 3,546 & 685 & 2,861 & 19.32 & 6.86 & 0.340 \\
Baseline & 5120 & 3,598 & 895 & 2,703 & 24.88 & 6.84 & 0.350 \\
EvoLen & 5120 & 4,290 & 833 & 3,457 & 19.42 & 7.03 & 0.342 \\
\bottomrule
\end{tabular}
\end{table}

\begin{table}[H]
\centering
\small
\caption{Exon versus intron token enrichment summary.}\label{tab:S5_exon_intron_summary}
\begin{tabular}{@{}llccccc@{}}
\toprule
Model & Vocab & SigTokens & SigExon & SigIntron & BestExon & BestIntron \\
\midrule
Baseline & 2048 & 1,350 & 470 & 880 & A & TATT \\
EvoLen & 2048 & 1,532 & 558 & 974 & C & T \\
Baseline & 3072 & 1,754 & 603 & 1,151 & A & TATT \\
EvoLen & 3072 & 2,026 & 720 & 1,306 & C & T \\
Baseline & 4096 & 2,066 & 714 & 1,352 & A & TATT \\
EvoLen & 4096 & 2,442 & 857 & 1,585 & C & T \\
Baseline & 5120 & 2,266 & 786 & 1,480 & A & TATT \\
EvoLen & 5120 & 2,776 & 1,025 & 1,751 & C & T \\
\bottomrule
\end{tabular}
\end{table}

\subsection{Ablation Study Details}

To isolate the contribution of each component, we design three ablation variants at vocabulary size 5,120: \textbf{No Partition} removes three-way genomic partitioning and trains a single whole-genome BPE; \textbf{No Priority} removes the conservation-priority merge order; \textbf{Linear score} replaces $\ell^2$ scoring with linear $\ell$. All other aspects remain identical to full EvoLen.

Table~\ref{tab:ablation_structural} reports structural metrics (see Table~\ref{tab:S1_motif_preservation} for metric definitions of Tok/Motif and Perfect\%) and Table~\ref{tab:ablation_downstream} reports downstream performance.

\begin{table}[htbp]
\centering
\caption{Ablation structural analysis (vocab size 5,120). \textbf{Bold}: best ablation per metric.}\label{tab:ablation_structural}
\small
\begin{tabular}{@{}lccccc@{}}
\toprule
Tokenizer & Mean phyloP & \% $>$ 0 & Mean Var & Tok/Motif & Perfect\% \\
\midrule
\multicolumn{6}{l}{\textit{References}} \\
Baseline BPE        & 0.082 & 85.0 & 0.357 & 2.78 & 1.66 \\
EvoLen        & 0.097 & 88.7 & 0.361 & 2.80 & 2.11 \\
\midrule
\multicolumn{6}{l}{\textit{Ablations}} \\
No Partition        & 0.085 & 85.9 & \textbf{0.359} & \textbf{2.78} & 1.66 \\
No Priority         & 0.071 & 81.0 & 0.374 & 3.08 & 1.69 \\
Linear score ($\ell$ vs $\ell^2$) & \textbf{0.100} & \textbf{88.7} & \textbf{0.359} & 2.86 & \textbf{2.11} \\
\bottomrule
\end{tabular}

\vspace{0.5em}

\small
\begin{tabular}{@{}llcccc@{}}
\toprule
Tokenizer & Region & Mean phyloP & \% $>$ 0 & Mean Var & Tokens \\
\midrule
No Partition  & conserved   &  0.547 & 100.0 & 0.291 & 5,060 \\
No Partition  & neutral     &  0.072 &  85.2 & 0.357 & 5,109 \\
No Partition  & accelerated & $-$0.273 &   0.5 & 0.652 & 5,109 \\
\addlinespace
No Priority   & conserved   &  0.539 &  99.9 & 0.293 & 5,006 \\
No Priority   & neutral     &  0.060 &  80.2 & 0.367 & 5,111 \\
No Priority   & accelerated & $-$0.276 &   0.5 & 0.661 & 5,111 \\
\addlinespace
Linear score     & conserved   &  0.558 & 100.0 & 0.289 & 5,104 \\
Linear score     & neutral     &  0.083 &  87.6 & 0.357 & 5,114 \\
Linear score     & accelerated & $-$0.270 &   0.2 & 0.652 & 5,114 \\
\addlinespace
EvoLen  & conserved   &  0.558 & 100.0 & 0.291 & 5,107 \\
EvoLen  & neutral     &  0.081 &  87.2 & 0.358 & 5,114 \\
EvoLen  & accelerated & $-$0.273 &   0.2 & 0.654 & 5,114 \\
\bottomrule
\end{tabular}
\end{table}

\vspace{0.5em}

\begin{table}[htbp]
\centering
\caption{Ablation downstream performance: average MCC (\%) comparing full EvoLen against variants without length encoding (w/o Len), without partition (w/o Part), and without priority (w/o Pri). \textbf{Bold} = best; \underline{underline} = second best.}\label{tab:ablation_downstream}
\small
\begin{tabular}{@{}lccccc@{}}
\toprule
Benchmark & Task Group & EvoLen & w/o Len & w/o Part & w/o Pri \\
\midrule
GUE & EMP (Yeast) & \textbf{46.8} & 45.5 & \underline{45.7} & 43.7 \\
GUE & Mouse & \textbf{58.8} & 50.2 & \underline{55.8} & 51.5 \\
GUE & Promoter-300 & \textbf{78.5} & 76.7 & 75.0 & \underline{77.3} \\
GUE & Promoter-Core & \textbf{63.6} & 61.1 & \underline{63.1} & 59.2 \\
GUE & Splice & \textbf{75.2} & \underline{71.6} & 69.0 & 68.9 \\
GUE & TF binding & \textbf{61.0} & 58.5 & \underline{59.0} & 54.5 \\
\midrule
GenomicBench & Human regulatory & \textbf{70.3} & 68.7 & \underline{68.8} & 67.9 \\
GenomicBench & Mouse enhancers & \textbf{63.8} & 46.3 & 47.0 & \underline{59.2} \\
GenomicBench & Invertebrates & \textbf{76.1} & \underline{74.6} & 74.0 & 71.1 \\
\midrule
NT & Histone marks & \textbf{55.0} & \underline{54.2} & 53.7 & 52.3 \\
NT & Enhancers & \textbf{49.6} & \underline{47.4} & 47.0 & 45.4 \\
NT & Promoters & \textbf{76.5} & \underline{71.9} & 69.8 & 70.6 \\
NT & Splice & \textbf{67.8} & \underline{66.3} & 61.2 & 58.2 \\
\midrule
Multi-SCREEN & Human cCRE & \textbf{22.1} & 21.6 & \underline{21.6} & 21.2 \\
\midrule
snATAC-seq & Human/Mouse brain & \underline{13.3} & 13.0 & \textbf{13.4} & 12.4 \\
\bottomrule
\end{tabular}
\end{table}

\subsection{Per-Task Downstream Results}\label{sec:per_task_results}

Tables~\ref{tab:per_task_mcc_baseline}--\ref{tab:per_task_recall_baseline} report per-task baseline vs.\ EvoLen results across all 56 downstream tasks, and Tables~\ref{tab:per_task_mcc_tokenizer}--\ref{tab:per_task_recall_ablation} report tokenizer-strategy and ablation comparisons. We separate comparisons for readability: baseline vs.\ EvoLen, tokenizer strategies (DNAbert2, NT, Grover), and ablation variants (w/o Len, w/o Part, w/o Pri). We report Matthews correlation coefficient (MCC), accuracy (Acc), macro F1, precision, and recall. The main-text Table~1 reports group-averaged MCC; these tables provide the full task-level breakdown.

\begin{table}[H]
\centering
\scriptsize
\setlength{\tabcolsep}{3pt}
\renewcommand{\arraystretch}{0.95}
\begin{tabular}{rllcc}
\toprule
\# & Benchmark & Task & Base & EvoLen \\
\midrule
1 & GBM & demo\_coding\_vs\_intergenomic\_seqs & 82.2 & 81.4 \\
2 & GBM & demo\_human\_or\_worm & 91.8 & 92.1 \\
3 & GBM & drosophila\_enhancers\_stark & 55.3 & 54.9 \\
4 & GBM & dummy\_mouse\_enhancers\_ensembl & 58.1 & 63.8 \\
5 & GBM & human\_enhancers\_cohn & 50.2 & 49.7 \\
6 & GBM & human\_enhancers\_ensembl & 77.5 & 78.3 \\
7 & GBM & human\_ensembl\_regulatory & 84.8 & 84.3 \\
8 & GBM & human\_nontata\_promoters & 87.4 & 89.1 \\
9 & GBM & human\_ocr\_ensembl & 49.9 & 50.0 \\
\midrule
10 & GUE & H3 & 69.6 & 73.9 \\
11 & GUE & H3K14ac & 40.7 & 40.1 \\
12 & GUE & H3K36me3 & 42.8 & 44.5 \\
13 & GUE & H3K4me1 & 41.2 & 36.3 \\
14 & GUE & H3K4me2 & 30.5 & 30.4 \\
15 & GUE & H3K4me3 & 25.5 & 26.1 \\
16 & GUE & H3K79me3 & 58.1 & 59.0 \\
17 & GUE & H3K9ac & 49.1 & 49.3 \\
18 & GUE & H4 & 72.7 & 73.9 \\
19 & GUE & H4ac & 37.6 & 34.6 \\
20 & GUE & mouse\_0 & 40.0 & 43.1 \\
21 & GUE & mouse\_1 & 79.7 & 79.1 \\
22 & GUE & mouse\_2 & 74.6 & 75.7 \\
23 & GUE & mouse\_3 & 55.6 & 60.7 \\
24 & GUE & mouse\_4 & 32.4 & 35.2 \\
25 & GUE & 300\_all & 82.5 & 82.7 \\
26 & GUE & 300\_notata & 89.5 & 90.7 \\
27 & GUE & 300\_tata & 57.4 & 62.1 \\
28 & GUE & core\_all & 62.3 & 62.7 \\
29 & GUE & core\_notata & 65.8 & 66.0 \\
30 & GUE & core\_tata & 64.6 & 62.2 \\
31 & GUE & splice\_reconstructed & 78.5 & 75.2 \\
32 & GUE & tf\_0 & 67.2 & 65.5 \\
33 & GUE & tf\_1 & 64.9 & 67.8 \\
34 & GUE & tf\_2 & 52.1 & 57.2 \\
35 & GUE & tf\_3 & 40.2 & 45.7 \\
36 & GUE & tf\_4 & 64.2 & 69.1 \\
\midrule
37 & snATAC-seq & Human/Mouse brain & 12.2 & 13.3 \\
\midrule
38 & Multi-SCREEN & Human cCRE & 21.9 & 22.1 \\
\midrule
39 & NT & H2AFZ & 48.1 & 49.3 \\
40 & NT & H3K27ac & 46.5 & 46.0 \\
41 & NT & H3K27me3 & 58.1 & 58.6 \\
42 & NT & H3K36me3 & 62.1 & 61.7 \\
43 & NT & H3K4me1 & 45.2 & 47.8 \\
44 & NT & H3K4me2 & 54.7 & 58.5 \\
45 & NT & H3K4me3 & 64.7 & 65.1 \\
46 & NT & H3K9ac & 53.8 & 54.6 \\
47 & NT & H3K9me3 & 38.9 & 42.0 \\
48 & NT & H4K20me1 & 65.2 & 66.0 \\
49 & NT & enhancers & 51.1 & 52.8 \\
50 & NT & enhancers\_types & 46.3 & 46.5 \\
51 & NT & promoter\_all & 73.9 & 74.4 \\
52 & NT & promoter\_no\_tata & 76.2 & 77.5 \\
53 & NT & promoter\_tata & 73.9 & 77.7 \\
54 & NT & splice\_sites\_acceptors & 66.9 & 66.6 \\
55 & NT & splice\_sites\_all & 67.4 & 68.0 \\
56 & NT & splice\_sites\_donors & 73.6 & 68.9 \\
\bottomrule
\end{tabular}
\caption{Per-task MCC (\%) for baseline vs EvoLen comparison.}
\label{tab:per_task_mcc_baseline}
\end{table}

\begin{table}[H]
\centering
\scriptsize
\setlength{\tabcolsep}{3pt}
\renewcommand{\arraystretch}{0.95}
\begin{tabular}{rllcc}
\toprule
\# & Benchmark & Task & Base & EvoLen \\
\midrule
1 & GBM & demo\_coding\_vs\_intergenomic\_seqs & 91.1 & 90.7 \\
2 & GBM & demo\_human\_or\_worm & 95.9 & 96.1 \\
3 & GBM & drosophila\_enhancers\_stark & 77.8 & 77.6 \\
4 & GBM & dummy\_mouse\_enhancers\_ensembl & 79.3 & 81.8 \\
5 & GBM & human\_enhancers\_cohn & 75.1 & 74.9 \\
6 & GBM & human\_enhancers\_ensembl & 88.7 & 89.1 \\
7 & GBM & human\_ensembl\_regulatory & 89.8 & 89.5 \\
8 & GBM & human\_nontata\_promoters & 93.5 & 94.5 \\
9 & GBM & human\_ocr\_ensembl & 74.9 & 75.0 \\
\midrule
10 & GUE & H3 & 84.7 & 87.0 \\
11 & GUE & H3K14ac & 71.0 & 70.6 \\
12 & GUE & H3K36me3 & 71.4 & 72.6 \\
13 & GUE & H3K4me1 & 71.0 & 68.3 \\
14 & GUE & H3K4me2 & 67.2 & 67.1 \\
15 & GUE & H3K4me3 & 62.9 & 63.1 \\
16 & GUE & H3K79me3 & 79.1 & 79.5 \\
17 & GUE & H3K9ac & 74.8 & 74.2 \\
18 & GUE & H4 & 86.5 & 87.1 \\
19 & GUE & H4ac & 68.9 & 67.4 \\
20 & GUE & mouse\_0 & 70.0 & 71.5 \\
21 & GUE & mouse\_1 & 89.8 & 89.5 \\
22 & GUE & mouse\_2 & 87.2 & 87.8 \\
23 & GUE & mouse\_3 & 77.8 & 80.3 \\
24 & GUE & mouse\_4 & 66.2 & 67.6 \\
25 & GUE & 300\_all & 91.2 & 91.4 \\
26 & GUE & 300\_notata & 94.7 & 95.3 \\
27 & GUE & 300\_tata & 78.6 & 81.1 \\
28 & GUE & core\_all & 81.2 & 81.3 \\
29 & GUE & core\_notata & 82.9 & 83.0 \\
30 & GUE & core\_tata & 82.1 & 81.1 \\
31 & GUE & splice\_reconstructed & 87.4 & 85.5 \\
32 & GUE & tf\_0 & 83.4 & 82.3 \\
33 & GUE & tf\_1 & 81.0 & 83.6 \\
34 & GUE & tf\_2 & 75.8 & 78.5 \\
35 & GUE & tf\_3 & 69.1 & 72.6 \\
36 & GUE & tf\_4 & 82.1 & 84.5 \\
\midrule
37 & snATAC-seq & Human/Mouse brain & 38.2 & 39.1 \\
\midrule
38 & Multi-SCREEN & Human cCRE & 31.5 & 31.6 \\
\midrule
39 & NT & H2AFZ & 73.9 & 74.1 \\
40 & NT & H3K27ac & 73.3 & 72.9 \\
41 & NT & H3K27me3 & 78.9 & 79.1 \\
42 & NT & H3K36me3 & 81.0 & 80.3 \\
43 & NT & H3K4me1 & 72.3 & 73.0 \\
44 & NT & H3K4me2 & 77.4 & 79.0 \\
45 & NT & H3K4me3 & 82.1 & 82.4 \\
46 & NT & H3K9ac & 76.9 & 77.2 \\
47 & NT & H3K9me3 & 69.4 & 71.0 \\
48 & NT & H4K20me1 & 82.5 & 83.0 \\
49 & NT & enhancers & 75.5 & 75.9 \\
50 & NT & enhancers\_types & 71.7 & 70.9 \\
51 & NT & promoter\_all & 86.9 & 87.2 \\
52 & NT & promoter\_no\_tata & 88.0 & 88.7 \\
53 & NT & promoter\_tata & 86.9 & 88.8 \\
54 & NT & splice\_sites\_acceptors & 83.4 & 83.3 \\
55 & NT & splice\_sites\_all & 78.3 & 78.6 \\
56 & NT & splice\_sites\_donors & 86.8 & 84.4 \\
\bottomrule
\end{tabular}
\caption{Per-task ACC (\%) for baseline vs EvoLen comparison.}
\label{tab:per_task_acc_baseline}
\end{table}

\begin{table}[H]
\centering
\scriptsize
\setlength{\tabcolsep}{3pt}
\renewcommand{\arraystretch}{0.95}
\begin{tabular}{rllcc}
\toprule
\# & Benchmark & Task & Base & EvoLen \\
\midrule
1 & GBM & demo\_coding\_vs\_intergenomic\_seqs & 91.1 & 90.7 \\
2 & GBM & demo\_human\_or\_worm & 95.9 & 96.1 \\
3 & GBM & drosophila\_enhancers\_stark & 77.3 & 77.4 \\
4 & GBM & dummy\_mouse\_enhancers\_ensembl & 79.1 & 81.7 \\
5 & GBM & human\_enhancers\_cohn & 75.1 & 74.9 \\
6 & GBM & human\_enhancers\_ensembl & 88.6 & 89.0 \\
7 & GBM & human\_ensembl\_regulatory & 90.0 & 89.6 \\
8 & GBM & human\_nontata\_promoters & 93.5 & 94.5 \\
9 & GBM & human\_ocr\_ensembl & 74.9 & 75.0 \\
\midrule
10 & GUE & H3 & 84.7 & 87.0 \\
11 & GUE & H3K14ac & 70.3 & 70.0 \\
12 & GUE & H3K36me3 & 71.3 & 72.2 \\
13 & GUE & H3K4me1 & 70.5 & 68.1 \\
14 & GUE & H3K4me2 & 64.0 & 64.5 \\
15 & GUE & H3K4me3 & 62.8 & 63.0 \\
16 & GUE & H3K79me3 & 79.0 & 79.4 \\
17 & GUE & H3K9ac & 74.6 & 74.2 \\
18 & GUE & H4 & 86.0 & 86.9 \\
19 & GUE & H4ac & 68.8 & 67.3 \\
20 & GUE & mouse\_0 & 70.0 & 71.4 \\
21 & GUE & mouse\_1 & 89.8 & 89.5 \\
22 & GUE & mouse\_2 & 87.2 & 87.8 \\
23 & GUE & mouse\_3 & 77.8 & 80.3 \\
24 & GUE & mouse\_4 & 66.1 & 67.6 \\
25 & GUE & 300\_all & 91.2 & 91.3 \\
26 & GUE & 300\_notata & 94.7 & 95.3 \\
27 & GUE & 300\_tata & 78.6 & 81.0 \\
28 & GUE & core\_all & 81.2 & 81.3 \\
29 & GUE & core\_notata & 82.9 & 83.0 \\
30 & GUE & core\_tata & 82.0 & 81.1 \\
31 & GUE & splice\_reconstructed & 86.6 & 84.5 \\
32 & GUE & tf\_0 & 83.3 & 82.2 \\
33 & GUE & tf\_1 & 80.6 & 83.5 \\
34 & GUE & tf\_2 & 75.7 & 78.5 \\
35 & GUE & tf\_3 & 68.3 & 72.4 \\
36 & GUE & tf\_4 & 82.1 & 84.5 \\
\midrule
37 & snATAC-seq & Human/Mouse brain & 27.1 & 32.0 \\
\midrule
38 & Multi-SCREEN & Human cCRE & 30.4 & 30.6 \\
\midrule
39 & NT & H2AFZ & 73.8 & 73.7 \\
40 & NT & H3K27ac & 73.3 & 72.8 \\
41 & NT & H3K27me3 & 78.9 & 79.1 \\
42 & NT & H3K36me3 & 81.0 & 80.2 \\
43 & NT & H3K4me1 & 72.2 & 72.7 \\
44 & NT & H3K4me2 & 77.3 & 78.7 \\
45 & NT & H3K4me3 & 82.0 & 82.4 \\
46 & NT & H3K9ac & 76.9 & 77.2 \\
47 & NT & H3K9me3 & 69.4 & 71.0 \\
48 & NT & H4K20me1 & 82.5 & 83.0 \\
49 & NT & enhancers & 75.5 & 75.7 \\
50 & NT & enhancers\_types & 48.8 & 55.4 \\
51 & NT & promoter\_all & 86.9 & 87.1 \\
52 & NT & promoter\_no\_tata & 88.0 & 88.7 \\
53 & NT & promoter\_tata & 86.9 & 88.8 \\
54 & NT & splice\_sites\_acceptors & 83.4 & 83.3 \\
55 & NT & splice\_sites\_all & 78.3 & 78.6 \\
56 & NT & splice\_sites\_donors & 86.8 & 84.4 \\
\bottomrule
\end{tabular}
\caption{Per-task F1 (\%) for baseline vs EvoLen comparison.}
\label{tab:per_task_f1_baseline}
\end{table}

\begin{table}[H]
\centering
\scriptsize
\setlength{\tabcolsep}{3pt}
\renewcommand{\arraystretch}{0.95}
\begin{tabular}{rllcc}
\toprule
\# & Benchmark & Task & Base & EvoLen \\
\midrule
1 & GBM & demo\_coding\_vs\_intergenomic\_seqs & 91.1 & 90.7 \\
2 & GBM & demo\_human\_or\_worm & 95.9 & 96.1 \\
3 & GBM & drosophila\_enhancers\_stark & 78.2 & 77.5 \\
4 & GBM & dummy\_mouse\_enhancers\_ensembl & 79.0 & 81.7 \\
5 & GBM & human\_enhancers\_cohn & 75.1 & 74.9 \\
6 & GBM & human\_enhancers\_ensembl & 88.9 & 89.3 \\
7 & GBM & human\_ensembl\_regulatory & 90.4 & 90.2 \\
8 & GBM & human\_nontata\_promoters & 93.5 & 94.4 \\
9 & GBM & human\_ocr\_ensembl & 74.9 & 75.0 \\
\midrule
10 & GUE & H3 & 84.9 & 87.0 \\
11 & GUE & H3K14ac & 70.4 & 70.0 \\
12 & GUE & H3K36me3 & 71.3 & 72.2 \\
13 & GUE & H3K4me1 & 70.8 & 68.1 \\
14 & GUE & H3K4me2 & 66.6 & 66.1 \\
15 & GUE & H3K4me3 & 62.8 & 63.0 \\
16 & GUE & H3K79me3 & 79.1 & 79.4 \\
17 & GUE & H3K9ac & 74.5 & 74.5 \\
18 & GUE & H4 & 87.1 & 87.1 \\
19 & GUE & H4ac & 68.8 & 67.3 \\
20 & GUE & mouse\_0 & 70.0 & 71.6 \\
21 & GUE & mouse\_1 & 89.8 & 89.5 \\
22 & GUE & mouse\_2 & 87.4 & 87.9 \\
23 & GUE & mouse\_3 & 77.8 & 80.4 \\
24 & GUE & mouse\_4 & 66.2 & 67.6 \\
25 & GUE & 300\_all & 91.2 & 91.4 \\
26 & GUE & 300\_notata & 94.8 & 95.3 \\
27 & GUE & 300\_tata & 78.7 & 81.1 \\
28 & GUE & core\_all & 81.2 & 81.4 \\
29 & GUE & core\_notata & 82.9 & 83.0 \\
30 & GUE & core\_tata & 82.5 & 81.1 \\
31 & GUE & splice\_reconstructed & 86.2 & 84.6 \\
32 & GUE & tf\_0 & 83.9 & 83.2 \\
33 & GUE & tf\_1 & 84.0 & 84.2 \\
34 & GUE & tf\_2 & 76.3 & 78.7 \\
35 & GUE & tf\_3 & 71.2 & 73.1 \\
36 & GUE & tf\_4 & 82.2 & 84.6 \\
\midrule
37 & snATAC-seq & Human/Mouse brain & 36.4 & 35.4 \\
\midrule
38 & Multi-SCREEN & Human cCRE & 30.7 & 31.2 \\
\midrule
39 & NT & H2AFZ & 74.3 & 75.4 \\
40 & NT & H3K27ac & 73.3 & 73.1 \\
41 & NT & H3K27me3 & 79.2 & 79.4 \\
42 & NT & H3K36me3 & 81.1 & 81.3 \\
43 & NT & H3K4me1 & 72.8 & 74.6 \\
44 & NT & H3K4me2 & 77.4 & 79.8 \\
45 & NT & H3K4me3 & 82.6 & 82.7 \\
46 & NT & H3K9ac & 76.9 & 77.3 \\
47 & NT & H3K9me3 & 69.5 & 71.0 \\
48 & NT & H4K20me1 & 82.7 & 83.0 \\
49 & NT & enhancers & 75.5 & 76.7 \\
50 & NT & enhancers\_types & 48.0 & 65.4 \\
51 & NT & promoter\_all & 86.9 & 87.2 \\
52 & NT & promoter\_no\_tata & 88.3 & 88.8 \\
53 & NT & promoter\_tata & 86.9 & 88.9 \\
54 & NT & splice\_sites\_acceptors & 83.4 & 83.3 \\
55 & NT & splice\_sites\_all & 78.3 & 78.6 \\
56 & NT & splice\_sites\_donors & 86.8 & 84.5 \\
\bottomrule
\end{tabular}
\caption{Per-task Precision (\%) for baseline vs EvoLen comparison.}
\label{tab:per_task_precision_baseline}
\end{table}

\begin{table}[H]
\centering
\scriptsize
\setlength{\tabcolsep}{3pt}
\renewcommand{\arraystretch}{0.95}
\begin{tabular}{rllcc}
\toprule
\# & Benchmark & Task & Base & EvoLen \\
\midrule
1 & GBM & demo\_coding\_vs\_intergenomic\_seqs & 91.1 & 90.7 \\
2 & GBM & demo\_human\_or\_worm & 95.9 & 96.1 \\
3 & GBM & drosophila\_enhancers\_stark & 77.1 & 77.3 \\
4 & GBM & dummy\_mouse\_enhancers\_ensembl & 79.1 & 82.2 \\
5 & GBM & human\_enhancers\_cohn & 75.1 & 74.9 \\
6 & GBM & human\_enhancers\_ensembl & 88.6 & 89.0 \\
7 & GBM & human\_ensembl\_regulatory & 89.7 & 89.3 \\
8 & GBM & human\_nontata\_promoters & 93.9 & 94.7 \\
9 & GBM & human\_ocr\_ensembl & 74.9 & 75.0 \\
\midrule
10 & GUE & H3 & 84.7 & 87.0 \\
11 & GUE & H3K14ac & 70.3 & 70.2 \\
12 & GUE & H3K36me3 & 71.5 & 72.3 \\
13 & GUE & H3K4me1 & 70.5 & 68.2 \\
14 & GUE & H3K4me2 & 64.0 & 64.3 \\
15 & GUE & H3K4me3 & 62.7 & 63.1 \\
16 & GUE & H3K79me3 & 79.0 & 79.5 \\
17 & GUE & H3K9ac & 74.6 & 74.7 \\
18 & GUE & H4 & 85.6 & 86.8 \\
19 & GUE & H4ac & 68.8 & 67.3 \\
20 & GUE & mouse\_0 & 70.0 & 71.5 \\
21 & GUE & mouse\_1 & 89.8 & 89.5 \\
22 & GUE & mouse\_2 & 87.2 & 87.8 \\
23 & GUE & mouse\_3 & 77.8 & 80.3 \\
24 & GUE & mouse\_4 & 66.2 & 67.6 \\
25 & GUE & 300\_all & 91.2 & 91.3 \\
26 & GUE & 300\_notata & 94.7 & 95.3 \\
27 & GUE & 300\_tata & 78.7 & 81.0 \\
28 & GUE & core\_all & 81.2 & 81.3 \\
29 & GUE & core\_notata & 82.9 & 83.0 \\
30 & GUE & core\_tata & 82.1 & 81.1 \\
31 & GUE & splice\_reconstructed & 86.9 & 84.4 \\
32 & GUE & tf\_0 & 83.4 & 82.3 \\
33 & GUE & tf\_1 & 81.0 & 83.6 \\
34 & GUE & tf\_2 & 75.8 & 78.5 \\
35 & GUE & tf\_3 & 69.1 & 72.6 \\
36 & GUE & tf\_4 & 82.1 & 84.5 \\
\midrule
37 & snATAC-seq & Human/Mouse brain & 31.6 & 33.1 \\
\midrule
38 & Multi-SCREEN & Human cCRE & 31.6 & 31.7 \\
\midrule
39 & NT & H2AFZ & 73.8 & 73.9 \\
40 & NT & H3K27ac & 73.3 & 72.9 \\
41 & NT & H3K27me3 & 79.0 & 79.2 \\
42 & NT & H3K36me3 & 81.0 & 80.4 \\
43 & NT & H3K4me1 & 72.4 & 73.2 \\
44 & NT & H3K4me2 & 77.3 & 78.7 \\
45 & NT & H3K4me3 & 82.1 & 82.4 \\
46 & NT & H3K9ac & 76.9 & 77.2 \\
47 & NT & H3K9me3 & 69.4 & 71.0 \\
48 & NT & H4K20me1 & 82.6 & 83.0 \\
49 & NT & enhancers & 75.5 & 76.0 \\
50 & NT & enhancers\_types & 50.0 & 54.0 \\
51 & NT & promoter\_all & 86.9 & 87.1 \\
52 & NT & promoter\_no\_tata & 88.0 & 88.6 \\
53 & NT & promoter\_tata & 86.9 & 88.8 \\
54 & NT & splice\_sites\_acceptors & 83.5 & 83.3 \\
55 & NT & splice\_sites\_all & 78.4 & 78.7 \\
56 & NT & splice\_sites\_donors & 86.8 & 84.4 \\
\bottomrule
\end{tabular}
\caption{Per-task Recall (\%) for baseline vs EvoLen comparison.}
\label{tab:per_task_recall_baseline}
\end{table}

\begin{table}[H]
\centering
\scriptsize
\setlength{\tabcolsep}{3pt}
\renewcommand{\arraystretch}{0.95}
\begin{tabular}{rllccc}
\toprule
\# & Benchmark & Task & DNAbert2 & NT & Grover \\
\midrule
1 & GBM & demo\_coding\_vs\_intergenomic\_seqs & 80.5 & 77.9 & 83.1 \\
2 & GBM & demo\_human\_or\_worm & 91.6 & 87.7 & 92.8 \\
3 & GBM & drosophila\_enhancers\_stark & 37.5 & 37.9 & 38.4 \\
4 & GBM & dummy\_mouse\_enhancers\_ensembl & 48.6 & 50.5 & 47.7 \\
5 & GBM & human\_enhancers\_cohn & 45.9 & 43.5 & 47.4 \\
6 & GBM & human\_enhancers\_ensembl & 78.7 & 66.9 & 81.1 \\
7 & GBM & human\_ensembl\_regulatory & 84.9 & 86.3 & 85.5 \\
8 & GBM & human\_nontata\_promoters & 86.3 & 71.1 & 90.5 \\
9 & GBM & human\_ocr\_ensembl & 47.7 & 44.1 & 55.5 \\
\midrule
10 & GUE & H3 & 71.1 & 65.9 & 74.9 \\
11 & GUE & H3K14ac & 37.8 & 30.4 & 41.3 \\
12 & GUE & H3K36me3 & 42.6 & 36.3 & 44.6 \\
13 & GUE & H3K4me1 & 37.6 & 34.1 & 38.1 \\
14 & GUE & H3K4me2 & 30.2 & 29.5 & 30.2 \\
15 & GUE & H3K4me3 & 26.2 & 17.0 & 29.9 \\
16 & GUE & H3K79me3 & 58.1 & 52.0 & 59.1 \\
17 & GUE & H3K9ac & 49.3 & 44.2 & 51.4 \\
18 & GUE & H4 & 75.7 & 66.9 & 72.9 \\
19 & GUE & H4ac & 34.0 & 27.4 & 39.7 \\
20 & GUE & mouse\_0 & 32.8 & 30.7 & 43.7 \\
21 & GUE & mouse\_1 & 78.9 & 72.0 & 80.5 \\
22 & GUE & mouse\_2 & 75.0 & 62.3 & 72.7 \\
23 & GUE & mouse\_3 & 49.9 & 44.9 & 66.0 \\
24 & GUE & mouse\_4 & 31.2 & 21.2 & 41.1 \\
25 & GUE & prom\_300\_all & 81.0 & 84.0 & 82.5 \\
26 & GUE & prom\_300\_notata & 90.1 & 89.1 & 89.8 \\
27 & GUE & prom\_300\_tata & 57.3 & 64.1 & 56.7 \\
28 & GUE & prom\_core\_all & 61.0 & 60.5 & 61.4 \\
29 & GUE & prom\_core\_notata & 64.5 & 62.5 & 65.6 \\
30 & GUE & prom\_core\_tata & 51.8 & 67.4 & 59.9 \\
31 & GUE & splice\_reconstructed & 73.5 & 75.3 & 79.9 \\
32 & GUE & tf\_0 & 64.2 & 61.3 & 67.1 \\
33 & GUE & tf\_1 & 69.3 & 64.6 & 68.1 \\
34 & GUE & tf\_2 & 55.8 & 45.9 & 59.8 \\
35 & GUE & tf\_3 & 37.7 & 34.7 & 45.3 \\
36 & GUE & tf\_4 & 68.1 & 63.0 & 71.6 \\
\midrule
37 & snATAC-seq & Human/Mouse brain & 8.8 & 9.8 & 14.5 \\
\midrule
38 & Multi-SCREEN & Human cCRE & 22.0 & 21.3 & 23.9 \\
\midrule
39 & NT & H2AFZ & 47.0 & 47.9 & 47.2 \\
40 & NT & H3K27ac & 46.2 & 42.3 & 47.0 \\
41 & NT & H3K27me3 & 60.0 & 60.7 & 61.1 \\
42 & NT & H3K36me3 & 60.4 & 57.4 & 60.3 \\
43 & NT & H3K4me1 & 47.6 & 47.7 & 48.4 \\
44 & NT & H3K4me2 & 56.8 & 55.8 & 56.6 \\
45 & NT & H3K4me3 & 66.4 & 65.6 & 67.7 \\
46 & NT & H3K9ac & 53.4 & 47.6 & 54.5 \\
47 & NT & H3K9me3 & 44.1 & 36.3 & 42.6 \\
48 & NT & H4K20me1 & 62.8 & 60.9 & 62.9 \\
49 & NT & enhancers & 47.7 & 49.6 & 50.1 \\
50 & NT & enhancers\_types & 46.0 & 43.9 & 45.8 \\
51 & NT & promoter\_all & 73.4 & 71.8 & 73.9 \\
52 & NT & promoter\_no\_tata & 73.6 & 75.3 & 74.7 \\
53 & NT & promoter\_tata & 71.6 & 72.9 & 72.0 \\
54 & NT & splice\_sites\_acceptors & 71.2 & 93.1 & 74.9 \\
55 & NT & splice\_sites\_all & 69.5 & 93.6 & 80.8 \\
56 & NT & splice\_sites\_donors & 67.9 & 95.0 & 82.9 \\
\bottomrule
\end{tabular}
\caption{Per-task MCC (\%) for tokenizer-strategy comparison.}
\label{tab:per_task_mcc_tokenizer}
\end{table}

\begin{table}[H]
\centering
\scriptsize
\setlength{\tabcolsep}{3pt}
\renewcommand{\arraystretch}{0.95}
\begin{tabular}{rllccc}
\toprule
\# & Benchmark & Task & w/o Len & w/o Part & w/o Pri \\
\midrule
1 & GBM & demo\_coding\_vs\_intergenomic\_seqs & 80.9 & 81.1 & 80.2 \\
2 & GBM & demo\_human\_or\_worm & 90.5 & 91.7 & 91.2 \\
3 & GBM & drosophila\_enhancers\_stark & 52.4 & 49.2 & 41.8 \\
4 & GBM & dummy\_mouse\_enhancers\_ensembl & 46.3 & 47.0 & 59.2 \\
5 & GBM & human\_enhancers\_cohn & 48.7 & 46.5 & 47.3 \\
6 & GBM & human\_enhancers\_ensembl & 76.9 & 77.3 & 75.4 \\
7 & GBM & human\_ensembl\_regulatory & 84.4 & 83.5 & 83.4 \\
8 & GBM & human\_nontata\_promoters & 85.8 & 87.5 & 86.1 \\
9 & GBM & human\_ocr\_ensembl & 48.0 & 49.3 & 47.1 \\
\midrule
10 & GUE & H3 & 69.8 & 70.0 & 66.6 \\
11 & GUE & H3K14ac & 38.7 & 39.0 & 34.7 \\
12 & GUE & H3K36me3 & 42.3 & 42.8 & 40.7 \\
13 & GUE & H3K4me1 & 37.0 & 38.6 & 36.3 \\
14 & GUE & H3K4me2 & 30.9 & 29.2 & 28.3 \\
15 & GUE & H3K4me3 & 25.5 & 23.8 & 25.7 \\
16 & GUE & H3K79me3 & 57.3 & 59.8 & 55.3 \\
17 & GUE & H3K9ac & 46.4 & 47.2 & 44.7 \\
18 & GUE & H4 & 72.9 & 72.7 & 70.9 \\
19 & GUE & H4ac & 34.1 & 34.0 & 33.6 \\
20 & GUE & mouse\_0 & 33.3 & 36.2 & 28.6 \\
21 & GUE & mouse\_1 & 77.2 & 77.6 & 76.0 \\
22 & GUE & mouse\_2 & 64.1 & 65.9 & 68.4 \\
23 & GUE & mouse\_3 & 44.0 & 68.3 & 54.0 \\
24 & GUE & mouse\_4 & 32.5 & 31.1 & 30.4 \\
25 & GUE & prom\_300\_all & 82.3 & 82.4 & 81.5 \\
26 & GUE & prom\_300\_notata & 89.1 & 90.8 & 89.9 \\
27 & GUE & prom\_300\_tata & 58.8 & 51.9 & 60.5 \\
28 & GUE & prom\_core\_all & 61.9 & 61.7 & 60.0 \\
29 & GUE & prom\_core\_notata & 64.8 & 65.4 & 64.6 \\
30 & GUE & prom\_core\_tata & 56.4 & 62.1 & 53.0 \\
31 & GUE & splice\_reconstructed & 71.6 & 69.0 & 68.9 \\
32 & GUE & tf\_0 & 66.4 & 64.0 & 62.0 \\
33 & GUE & tf\_1 & 66.2 & 68.1 & 64.9 \\
34 & GUE & tf\_2 & 52.9 & 54.6 & 46.6 \\
35 & GUE & tf\_3 & 42.2 & 42.4 & 36.3 \\
36 & GUE & tf\_4 & 64.8 & 65.9 & 62.6 \\
\midrule
37 & snATAC-seq & Human/Mouse brain & 13.0 & 13.4 & 12.4 \\
\midrule
38 & Multi-SCREEN & Human cCRE & 21.6 & 21.6 & 21.2 \\
\midrule
39 & NT & H2AFZ & 47.7 & 45.7 & 44.3 \\
40 & NT & H3K27ac & 46.2 & 43.8 & 42.4 \\
41 & NT & H3K27me3 & 60.2 & 61.3 & 60.3 \\
42 & NT & H3K36me3 & 58.9 & 56.7 & 57.1 \\
43 & NT & H3K4me1 & 48.7 & 48.5 & 46.8 \\
44 & NT & H3K4me2 & 57.0 & 57.1 & 54.5 \\
45 & NT & H3K4me3 & 66.6 & 66.4 & 65.1 \\
46 & NT & H3K9ac & 52.9 & 53.0 & 49.0 \\
47 & NT & H3K9me3 & 40.7 & 41.6 & 41.4 \\
48 & NT & H4K20me1 & 62.6 & 63.3 & 62.3 \\
49 & NT & enhancers & 48.0 & 47.7 & 46.2 \\
50 & NT & enhancers\_types & 46.7 & 46.3 & 44.5 \\
51 & NT & promoter\_all & 72.9 & 71.2 & 69.9 \\
52 & NT & promoter\_no\_tata & 74.3 & 74.1 & 73.7 \\
53 & NT & promoter\_tata & 68.6 & 64.1 & 68.2 \\
54 & NT & splice\_sites\_acceptors & 67.3 & 63.1 & 62.0 \\
55 & NT & splice\_sites\_all & 64.6 & 59.3 & 55.5 \\
56 & NT & splice\_sites\_donors & 66.9 & 61.2 & 57.1 \\
\bottomrule
\end{tabular}
\caption{Per-task MCC (\%) for ablation variants.}
\label{tab:per_task_mcc_ablation}
\end{table}

\begin{table}[H]
\centering
\scriptsize
\setlength{\tabcolsep}{3pt}
\renewcommand{\arraystretch}{0.95}
\begin{tabular}{rllccc}
\toprule
\# & Benchmark & Task & DNAbert2 & NT & Grover \\
\midrule
1 & GBM & demo\_coding\_vs\_intergenomic\_seqs & 90.2 & 89.0 & 91.5 \\
2 & GBM & demo\_human\_or\_worm & 95.8 & 93.8 & 96.4 \\
3 & GBM & drosophila\_enhancers\_stark & 69.1 & 69.1 & 69.5 \\
4 & GBM & dummy\_mouse\_enhancers\_ensembl & 71.9 & 75.2 & 72.7 \\
5 & GBM & human\_enhancers\_cohn & 72.8 & 71.7 & 73.6 \\
6 & GBM & human\_enhancers\_ensembl & 89.3 & 83.4 & 90.5 \\
7 & GBM & human\_ensembl\_regulatory & 89.9 & 90.8 & 90.3 \\
8 & GBM & human\_nontata\_promoters & 93.0 & 85.7 & 95.1 \\
9 & GBM & human\_ocr\_ensembl & 73.8 & 72.0 & 77.7 \\
\midrule
10 & GUE & H3 & 85.5 & 82.9 & 87.4 \\
11 & GUE & H3K14ac & 69.7 & 66.5 & 71.5 \\
12 & GUE & H3K36me3 & 71.6 & 68.4 & 72.5 \\
13 & GUE & H3K4me1 & 69.2 & 67.6 & 69.4 \\
14 & GUE & H3K4me2 & 67.1 & 66.7 & 67.1 \\
15 & GUE & H3K4me3 & 63.3 & 59.0 & 65.0 \\
16 & GUE & H3K79me3 & 79.1 & 76.1 & 79.6 \\
17 & GUE & H3K9ac & 74.7 & 72.1 & 76.0 \\
18 & GUE & H4 & 88.0 & 83.7 & 86.6 \\
19 & GUE & H4ac & 67.0 & 64.0 & 70.1 \\
20 & GUE & mouse\_0 & 66.4 & 65.3 & 71.9 \\
21 & GUE & mouse\_1 & 89.4 & 85.9 & 90.2 \\
22 & GUE & mouse\_2 & 87.5 & 81.1 & 86.3 \\
23 & GUE & mouse\_3 & 74.9 & 72.4 & 82.8 \\
24 & GUE & mouse\_4 & 65.6 & 60.6 & 70.5 \\
25 & GUE & prom\_300\_all & 90.5 & 92.0 & 91.2 \\
26 & GUE & prom\_300\_notata & 95.1 & 94.6 & 94.9 \\
27 & GUE & prom\_300\_tata & 78.6 & 82.1 & 78.3 \\
28 & GUE & prom\_core\_all & 80.5 & 80.2 & 80.7 \\
29 & GUE & prom\_core\_notata & 82.2 & 81.2 & 82.8 \\
30 & GUE & prom\_core\_tata & 75.9 & 83.7 & 79.9 \\
31 & GUE & splice\_reconstructed & 84.3 & 85.1 & 88.2 \\
32 & GUE & tf\_0 & 81.7 & 80.3 & 83.2 \\
33 & GUE & tf\_1 & 84.5 & 81.9 & 83.7 \\
34 & GUE & tf\_2 & 77.5 & 72.7 & 79.0 \\
35 & GUE & tf\_3 & 68.2 & 67.1 & 72.2 \\
36 & GUE & tf\_4 & 84.0 & 81.4 & 85.8 \\
\midrule
37 & snATAC-seq & Human/Mouse brain & 35.3 & 37.2 & 39.6 \\
\midrule
38 & Multi-SCREEN & Human cCRE & 31.4 & 30.9 & 33.2 \\
\midrule
39 & NT & H2AFZ & 73.5 & 73.4 & 73.3 \\
40 & NT & H3K27ac & 73.1 & 71.0 & 73.3 \\
41 & NT & H3K27me3 & 80.0 & 80.0 & 80.4 \\
42 & NT & H3K36me3 & 80.0 & 78.5 & 80.1 \\
43 & NT & H3K4me1 & 73.7 & 72.8 & 73.8 \\
44 & NT & H3K4me2 & 78.3 & 77.5 & 78.2 \\
45 & NT & H3K4me3 & 83.1 & 82.8 & 83.5 \\
46 & NT & H3K9ac & 76.7 & 73.7 & 77.2 \\
47 & NT & H3K9me3 & 72.0 & 68.1 & 71.3 \\
48 & NT & H4K20me1 & 81.4 & 80.4 & 81.4 \\
49 & NT & enhancers & 73.8 & 74.6 & 75.1 \\
50 & NT & enhancers\_types & 71.6 & 70.3 & 71.5 \\
51 & NT & promoter\_all & 86.7 & 85.9 & 86.9 \\
52 & NT & promoter\_no\_tata & 86.8 & 87.6 & 87.3 \\
53 & NT & promoter\_tata & 85.8 & 86.4 & 86.0 \\
54 & NT & splice\_sites\_acceptors & 85.6 & 96.5 & 87.4 \\
55 & NT & splice\_sites\_all & 79.7 & 95.8 & 87.2 \\
56 & NT & splice\_sites\_donors & 84.0 & 97.5 & 91.5 \\
\bottomrule
\end{tabular}
\caption{Per-task Acc (\%) for tokenizer-strategy comparison.}
\label{tab:per_task_acc_tokenizer}
\end{table}

\begin{table}[H]
\centering
\scriptsize
\setlength{\tabcolsep}{3pt}
\renewcommand{\arraystretch}{0.95}
\begin{tabular}{rllccc}
\toprule
\# & Benchmark & Task & w/o Len & w/o Part & w/o Pri \\
\midrule
1 & GBM & demo\_coding\_vs\_intergenomic\_seqs & 90.4 & 90.6 & 90.1 \\
2 & GBM & demo\_human\_or\_worm & 95.2 & 95.8 & 95.6 \\
3 & GBM & drosophila\_enhancers\_stark & 76.2 & 74.7 & 71.0 \\
4 & GBM & dummy\_mouse\_enhancers\_ensembl & 73.6 & 71.9 & 78.5 \\
5 & GBM & human\_enhancers\_cohn & 74.4 & 73.2 & 73.6 \\
6 & GBM & human\_enhancers\_ensembl & 88.4 & 88.6 & 87.6 \\
7 & GBM & human\_ensembl\_regulatory & 89.6 & 89.0 & 88.9 \\
8 & GBM & human\_nontata\_promoters & 92.7 & 93.6 & 92.9 \\
9 & GBM & human\_ocr\_ensembl & 73.9 & 74.6 & 73.5 \\
\midrule
10 & GUE & H3 & 84.8 & 84.9 & 83.3 \\
11 & GUE & H3K14ac & 70.3 & 69.6 & 67.7 \\
12 & GUE & H3K36me3 & 71.5 & 71.5 & 70.8 \\
13 & GUE & H3K4me1 & 69.0 & 69.7 & 68.6 \\
14 & GUE & H3K4me2 & 67.4 & 66.7 & 66.3 \\
15 & GUE & H3K4me3 & 63.0 & 62.2 & 63.2 \\
16 & GUE & H3K79me3 & 78.7 & 80.0 & 77.7 \\
17 & GUE & H3K9ac & 72.4 & 73.7 & 72.2 \\
18 & GUE & H4 & 86.7 & 86.5 & 85.7 \\
19 & GUE & H4ac & 67.3 & 67.3 & 67.0 \\
20 & GUE & mouse\_0 & 66.7 & 68.0 & 64.3 \\
21 & GUE & mouse\_1 & 88.6 & 88.8 & 88.0 \\
22 & GUE & mouse\_2 & 82.0 & 82.9 & 84.1 \\
23 & GUE & mouse\_3 & 72.0 & 84.1 & 77.0 \\
24 & GUE & mouse\_4 & 66.1 & 65.5 & 65.1 \\
25 & GUE & prom\_300\_all & 91.2 & 91.2 & 90.7 \\
26 & GUE & prom\_300\_notata & 94.5 & 95.4 & 94.9 \\
27 & GUE & prom\_300\_tata & 79.1 & 75.7 & 80.1 \\
28 & GUE & prom\_core\_all & 80.9 & 80.9 & 80.0 \\
29 & GUE & prom\_core\_notata & 82.4 & 82.7 & 82.3 \\
30 & GUE & prom\_core\_tata & 77.8 & 80.8 & 76.5 \\
31 & GUE & splice\_reconstructed & 83.4 & 81.9 & 81.9 \\
32 & GUE & tf\_0 & 83.0 & 81.8 & 80.7 \\
33 & GUE & tf\_1 & 82.6 & 83.7 & 82.0 \\
34 & GUE & tf\_2 & 75.8 & 77.3 & 73.0 \\
35 & GUE & tf\_3 & 71.0 & 70.4 & 67.5 \\
36 & GUE & tf\_4 & 82.4 & 82.9 & 81.3 \\
\midrule
37 & snATAC-seq & Human/Mouse brain & 38.3 & 38.8 & 38.3 \\
\midrule
38 & Multi-SCREEN & Human cCRE & 31.1 & 31.2 & 30.8 \\
\midrule
39 & NT & H2AFZ & 73.8 & 72.0 & 72.0 \\
40 & NT & H3K27ac & 72.8 & 71.7 & 71.0 \\
41 & NT & H3K27me3 & 80.0 & 80.6 & 79.5 \\
42 & NT & H3K36me3 & 79.3 & 78.3 & 78.5 \\
43 & NT & H3K4me1 & 74.2 & 73.9 & 72.9 \\
44 & NT & H3K4me2 & 78.4 & 78.4 & 77.0 \\
45 & NT & H3K4me3 & 83.3 & 83.2 & 82.6 \\
46 & NT & H3K9ac & 76.5 & 76.5 & 74.4 \\
47 & NT & H3K9me3 & 70.4 & 70.8 & 70.7 \\
48 & NT & H4K20me1 & 81.3 & 81.7 & 81.1 \\
49 & NT & enhancers & 74.0 & 73.9 & 73.1 \\
50 & NT & enhancers\_types & 71.9 & 71.8 & 70.9 \\
51 & NT & promoter\_all & 86.1 & 85.6 & 85.0 \\
52 & NT & promoter\_no\_tata & 87.1 & 87.0 & 86.7 \\
53 & NT & promoter\_tata & 84.3 & 82.0 & 84.1 \\
54 & NT & splice\_sites\_acceptors & 83.7 & 81.5 & 81.0 \\
55 & NT & splice\_sites\_all & 76.4 & 72.8 & 70.2 \\
56 & NT & splice\_sites\_donors & 83.5 & 80.6 & 78.5 \\
\bottomrule
\end{tabular}
\caption{Per-task Acc (\%) for ablation variants.}
\label{tab:per_task_acc_ablation}
\end{table}

\begin{table}[H]
\centering
\scriptsize
\setlength{\tabcolsep}{3pt}
\renewcommand{\arraystretch}{0.95}
\begin{tabular}{rllccc}
\toprule
\# & Benchmark & Task & DNAbert2 & NT & Grover \\
\midrule
1 & GBM & demo\_coding\_vs\_intergenomic\_seqs & 90.2 & 89.0 & 91.5 \\
2 & GBM & demo\_human\_or\_worm & 95.8 & 93.8 & 96.4 \\
3 & GBM & drosophila\_enhancers\_stark & 68.5 & 68.9 & 68.8 \\
4 & GBM & dummy\_mouse\_enhancers\_ensembl & 71.8 & 75.1 & 72.7 \\
5 & GBM & human\_enhancers\_cohn & 72.8 & 71.7 & 73.6 \\
6 & GBM & human\_enhancers\_ensembl & 89.3 & 83.4 & 90.5 \\
7 & GBM & human\_ensembl\_regulatory & 90.1 & 90.9 & 90.5 \\
8 & GBM & human\_nontata\_promoters & 93.0 & 85.5 & 95.1 \\
9 & GBM & human\_ocr\_ensembl & 73.8 & 71.9 & 77.7 \\
\midrule
10 & GUE & H3 & 85.5 & 82.9 & 87.4 \\
11 & GUE & H3K14ac & 68.8 & 64.9 & 70.6 \\
12 & GUE & H3K36me3 & 71.3 & 68.1 & 72.3 \\
13 & GUE & H3K4me1 & 68.8 & 66.8 & 69.0 \\
14 & GUE & H3K4me2 & 63.8 & 63.8 & 63.9 \\
15 & GUE & H3K4me3 & 63.1 & 57.7 & 64.9 \\
16 & GUE & H3K79me3 & 79.0 & 75.9 & 79.5 \\
17 & GUE & H3K9ac & 74.5 & 72.0 & 75.7 \\
18 & GUE & H4 & 87.8 & 83.4 & 86.2 \\
19 & GUE & H4ac & 66.9 & 63.4 & 69.8 \\
20 & GUE & mouse\_0 & 66.4 & 65.3 & 71.8 \\
21 & GUE & mouse\_1 & 89.4 & 85.9 & 90.2 \\
22 & GUE & mouse\_2 & 87.5 & 81.1 & 86.3 \\
23 & GUE & mouse\_3 & 74.8 & 72.4 & 82.8 \\
24 & GUE & mouse\_4 & 65.6 & 60.6 & 70.5 \\
25 & GUE & prom\_300\_all & 90.5 & 92.0 & 91.2 \\
26 & GUE & prom\_300\_notata & 95.1 & 94.6 & 94.9 \\
27 & GUE & prom\_300\_tata & 78.6 & 82.0 & 78.3 \\
28 & GUE & prom\_core\_all & 80.5 & 80.2 & 80.7 \\
29 & GUE & prom\_core\_notata & 82.2 & 81.2 & 82.8 \\
30 & GUE & prom\_core\_tata & 75.8 & 83.7 & 79.9 \\
31 & GUE & splice\_reconstructed & 83.4 & 84.3 & 87.4 \\
32 & GUE & tf\_0 & 81.6 & 80.2 & 83.1 \\
33 & GUE & tf\_1 & 84.5 & 81.8 & 83.6 \\
34 & GUE & tf\_2 & 77.3 & 72.5 & 78.7 \\
35 & GUE & tf\_3 & 67.7 & 66.8 & 71.9 \\
36 & GUE & tf\_4 & 84.0 & 81.4 & 85.8 \\
\midrule
37 & snATAC-seq & Human/Mouse brain & 28.6 & 28.8 & 32.0 \\
\midrule
38 & Multi-SCREEN & Human cCRE & 30.1 & 29.7 & 32.1 \\
\midrule
39 & NT & H2AFZ & 73.5 & 73.1 & 73.3 \\
40 & NT & H3K27ac & 73.1 & 71.0 & 73.3 \\
41 & NT & H3K27me3 & 80.0 & 79.9 & 80.4 \\
42 & NT & H3K36me3 & 80.0 & 78.4 & 80.1 \\
43 & NT & H3K4me1 & 73.7 & 72.4 & 73.7 \\
44 & NT & H3K4me2 & 78.2 & 77.2 & 78.2 \\
45 & NT & H3K4me3 & 83.1 & 82.8 & 83.5 \\
46 & NT & H3K9ac & 76.7 & 73.5 & 77.2 \\
47 & NT & H3K9me3 & 71.9 & 67.7 & 71.3 \\
48 & NT & H4K20me1 & 81.4 & 80.3 & 81.4 \\
49 & NT & enhancers & 73.8 & 74.5 & 75.0 \\
50 & NT & enhancers\_types & 50.6 & 47.8 & 49.2 \\
51 & NT & promoter\_all & 86.7 & 85.9 & 86.9 \\
52 & NT & promoter\_no\_tata & 86.8 & 87.6 & 87.3 \\
53 & NT & promoter\_tata & 85.8 & 86.4 & 86.0 \\
54 & NT & splice\_sites\_acceptors & 85.6 & 96.5 & 87.4 \\
55 & NT & splice\_sites\_all & 79.7 & 95.7 & 87.2 \\
56 & NT & splice\_sites\_donors & 84.0 & 97.5 & 91.4 \\
\bottomrule
\end{tabular}
\caption{Per-task F1 (\%) for tokenizer-strategy comparison.}
\label{tab:per_task_f1_tokenizer}
\end{table}

\begin{table}[H]
\centering
\scriptsize
\setlength{\tabcolsep}{3pt}
\renewcommand{\arraystretch}{0.95}
\begin{tabular}{rllccc}
\toprule
\# & Benchmark & Task & w/o Len & w/o Part & w/o Pri \\
\midrule
1 & GBM & demo\_coding\_vs\_intergenomic\_seqs & 90.4 & 90.6 & 90.1 \\
2 & GBM & demo\_human\_or\_worm & 95.2 & 95.8 & 95.6 \\
3 & GBM & drosophila\_enhancers\_stark & 76.1 & 74.6 & 70.9 \\
4 & GBM & dummy\_mouse\_enhancers\_ensembl & 73.1 & 71.9 & 78.5 \\
5 & GBM & human\_enhancers\_cohn & 74.4 & 73.2 & 73.6 \\
6 & GBM & human\_enhancers\_ensembl & 88.4 & 88.6 & 87.6 \\
7 & GBM & human\_ensembl\_regulatory & 89.7 & 89.1 & 89.0 \\
8 & GBM & human\_nontata\_promoters & 92.7 & 93.6 & 92.9 \\
9 & GBM & human\_ocr\_ensembl & 73.9 & 74.6 & 73.5 \\
\midrule
10 & GUE & H3 & 84.8 & 84.9 & 83.3 \\
11 & GUE & H3K14ac & 69.3 & 69.3 & 67.3 \\
12 & GUE & H3K36me3 & 71.2 & 71.3 & 70.3 \\
13 & GUE & H3K4me1 & 68.4 & 69.2 & 68.1 \\
14 & GUE & H3K4me2 & 64.1 & 63.0 & 62.5 \\
15 & GUE & H3K4me3 & 62.5 & 61.5 & 62.7 \\
16 & GUE & H3K79me3 & 78.7 & 79.9 & 77.6 \\
17 & GUE & H3K9ac & 72.4 & 73.5 & 72.1 \\
18 & GUE & H4 & 86.4 & 86.3 & 85.4 \\
19 & GUE & H4ac & 67.0 & 66.8 & 66.8 \\
20 & GUE & mouse\_0 & 66.7 & 68.0 & 64.3 \\
21 & GUE & mouse\_1 & 88.6 & 88.8 & 88.0 \\
22 & GUE & mouse\_2 & 82.0 & 82.9 & 84.1 \\
23 & GUE & mouse\_3 & 71.9 & 84.1 & 77.0 \\
24 & GUE & mouse\_4 & 66.0 & 65.4 & 65.0 \\
25 & GUE & prom\_300\_all & 91.2 & 91.2 & 90.7 \\
26 & GUE & prom\_300\_notata & 94.5 & 95.4 & 94.9 \\
27 & GUE & prom\_300\_tata & 79.1 & 75.7 & 80.1 \\
28 & GUE & prom\_core\_all & 80.9 & 80.9 & 80.0 \\
29 & GUE & prom\_core\_notata & 82.4 & 82.7 & 82.3 \\
30 & GUE & prom\_core\_tata & 77.7 & 80.7 & 76.5 \\
31 & GUE & splice\_reconstructed & 82.2 & 80.6 & 80.6 \\
32 & GUE & tf\_0 & 83.0 & 81.7 & 80.6 \\
33 & GUE & tf\_1 & 82.5 & 83.6 & 81.9 \\
34 & GUE & tf\_2 & 75.5 & 77.3 & 72.8 \\
35 & GUE & tf\_3 & 70.9 & 69.8 & 66.9 \\
36 & GUE & tf\_4 & 82.4 & 82.9 & 81.3 \\
\midrule
37 & snATAC-seq & Human/Mouse brain & 30.3 & 29.5 & 31.0 \\
\midrule
38 & Multi-SCREEN & Human cCRE & 30.1 & 30.3 & 29.7 \\
\midrule
39 & NT & H2AFZ & 73.7 & 71.7 & 72.0 \\
40 & NT & H3K27ac & 72.7 & 71.6 & 70.8 \\
41 & NT & H3K27me3 & 80.0 & 80.6 & 79.3 \\
42 & NT & H3K36me3 & 79.2 & 78.3 & 78.5 \\
43 & NT & H3K4me1 & 74.1 & 73.8 & 72.7 \\
44 & NT & H3K4me2 & 78.4 & 78.3 & 76.9 \\
45 & NT & H3K4me3 & 83.3 & 83.2 & 82.6 \\
46 & NT & H3K9ac & 76.4 & 76.5 & 74.3 \\
47 & NT & H3K9me3 & 70.3 & 70.7 & 70.7 \\
48 & NT & H4K20me1 & 81.3 & 81.6 & 81.1 \\
49 & NT & enhancers & 73.9 & 73.9 & 73.0 \\
50 & NT & enhancers\_types & 50.8 & 50.8 & 49.2 \\
51 & NT & promoter\_all & 86.1 & 85.6 & 85.0 \\
52 & NT & promoter\_no\_tata & 87.1 & 87.0 & 86.7 \\
53 & NT & promoter\_tata & 84.3 & 82.0 & 84.1 \\
54 & NT & splice\_sites\_acceptors & 83.7 & 81.5 & 81.0 \\
55 & NT & splice\_sites\_all & 76.4 & 72.8 & 70.2 \\
56 & NT & splice\_sites\_donors & 83.4 & 80.6 & 78.4 \\
\bottomrule
\end{tabular}
\caption{Per-task F1 (\%) for ablation variants.}
\label{tab:per_task_f1_ablation}
\end{table}

\begin{table}[H]
\centering
\scriptsize
\setlength{\tabcolsep}{3pt}
\renewcommand{\arraystretch}{0.95}
\begin{tabular}{rllccc}
\toprule
\# & Benchmark & Task & DNAbert2 & NT & Grover \\
\midrule
1 & GBM & demo\_coding\_vs\_intergenomic\_seqs & 90.3 & 89.0 & 91.5 \\
2 & GBM & demo\_human\_or\_worm & 95.8 & 93.8 & 96.4 \\
3 & GBM & drosophila\_enhancers\_stark & 69.1 & 68.9 & 69.7 \\
4 & GBM & dummy\_mouse\_enhancers\_ensembl & 74.8 & 75.1 & 73.9 \\
5 & GBM & human\_enhancers\_cohn & 73.1 & 71.8 & 73.8 \\
6 & GBM & human\_enhancers\_ensembl & 89.4 & 83.5 & 90.6 \\
7 & GBM & human\_ensembl\_regulatory & 90.6 & 91.5 & 90.8 \\
8 & GBM & human\_nontata\_promoters & 92.9 & 85.5 & 95.1 \\
9 & GBM & human\_ocr\_ensembl & 73.8 & 72.1 & 77.7 \\
\midrule
10 & GUE & H3 & 85.6 & 83.0 & 87.5 \\
11 & GUE & H3K14ac & 69.0 & 65.6 & 70.8 \\
12 & GUE & H3K36me3 & 71.3 & 68.1 & 72.2 \\
13 & GUE & H3K4me1 & 68.9 & 67.4 & 69.2 \\
14 & GUE & H3K4me2 & 66.5 & 65.8 & 66.4 \\
15 & GUE & H3K4me3 & 63.1 & 58.9 & 64.9 \\
16 & GUE & H3K79me3 & 79.1 & 76.2 & 79.5 \\
17 & GUE & H3K9ac & 74.5 & 72.0 & 75.7 \\
18 & GUE & H4 & 87.9 & 83.6 & 87.2 \\
19 & GUE & H4ac & 67.0 & 63.9 & 69.9 \\
20 & GUE & mouse\_0 & 66.4 & 65.4 & 71.9 \\
21 & GUE & mouse\_1 & 89.5 & 86.1 & 90.2 \\
22 & GUE & mouse\_2 & 87.5 & 81.2 & 86.4 \\
23 & GUE & mouse\_3 & 75.1 & 72.5 & 83.1 \\
24 & GUE & mouse\_4 & 65.6 & 60.6 & 70.6 \\
25 & GUE & prom\_300\_all & 90.5 & 92.0 & 91.3 \\
26 & GUE & prom\_300\_notata & 95.1 & 94.6 & 94.9 \\
27 & GUE & prom\_300\_tata & 78.6 & 82.0 & 78.3 \\
28 & GUE & prom\_core\_all & 80.5 & 80.3 & 80.7 \\
29 & GUE & prom\_core\_notata & 82.3 & 81.3 & 82.8 \\
30 & GUE & prom\_core\_tata & 75.9 & 83.7 & 79.9 \\
31 & GUE & splice\_reconstructed & 82.9 & 83.1 & 87.2 \\
32 & GUE & tf\_0 & 82.5 & 81.0 & 83.9 \\
33 & GUE & tf\_1 & 84.8 & 82.7 & 84.4 \\
34 & GUE & tf\_2 & 78.3 & 73.2 & 80.8 \\
35 & GUE & tf\_3 & 69.5 & 67.6 & 73.1 \\
36 & GUE & tf\_4 & 84.1 & 81.6 & 85.8 \\
\midrule
37 & snATAC-seq & Human/Mouse brain & 33.2 & 32.8 & 38.1 \\
\midrule
38 & Multi-SCREEN & Human cCRE & 30.7 & 30.3 & 32.6 \\
\midrule
39 & NT & H2AFZ & 73.5 & 74.4 & 73.7 \\
40 & NT & H3K27ac & 73.1 & 71.3 & 73.6 \\
41 & NT & H3K27me3 & 80.0 & 80.6 & 80.7 \\
42 & NT & H3K36me3 & 80.3 & 78.9 & 80.1 \\
43 & NT & H3K4me1 & 73.9 & 74.8 & 74.5 \\
44 & NT & H3K4me2 & 78.5 & 78.4 & 78.5 \\
45 & NT & H3K4me3 & 83.3 & 82.8 & 84.1 \\
46 & NT & H3K9ac & 76.7 & 74.1 & 77.2 \\
47 & NT & H3K9me3 & 72.1 & 68.5 & 71.3 \\
48 & NT & H4K20me1 & 81.4 & 80.5 & 81.5 \\
49 & NT & enhancers & 73.8 & 75.1 & 75.1 \\
50 & NT & enhancers\_types & 61.3 & 47.5 & 81.2 \\
51 & NT & promoter\_all & 86.7 & 85.9 & 87.0 \\
52 & NT & promoter\_no\_tata & 86.8 & 87.7 & 87.4 \\
53 & NT & promoter\_tata & 85.8 & 86.5 & 86.0 \\
54 & NT & splice\_sites\_acceptors & 85.6 & 96.6 & 87.5 \\
55 & NT & splice\_sites\_all & 79.7 & 95.7 & 87.2 \\
56 & NT & splice\_sites\_donors & 84.0 & 97.5 & 91.5 \\
\bottomrule
\end{tabular}
\caption{Per-task Precision (\%) for tokenizer-strategy comparison.}
\label{tab:per_task_prec_tokenizer}
\end{table}

\begin{table}[H]
\centering
\scriptsize
\setlength{\tabcolsep}{3pt}
\renewcommand{\arraystretch}{0.95}
\begin{tabular}{rllccc}
\toprule
\# & Benchmark & Task & w/o Len & w/o Part & w/o Pri \\
\midrule
1 & GBM & demo\_coding\_vs\_intergenomic\_seqs & 90.4 & 90.6 & 90.1 \\
2 & GBM & demo\_human\_or\_worm & 95.3 & 95.8 & 95.6 \\
3 & GBM & drosophila\_enhancers\_stark & 76.1 & 74.6 & 70.9 \\
4 & GBM & dummy\_mouse\_enhancers\_ensembl & 73.1 & 73.7 & 79.5 \\
5 & GBM & human\_enhancers\_cohn & 74.4 & 73.3 & 73.7 \\
6 & GBM & human\_enhancers\_ensembl & 88.5 & 88.8 & 87.7 \\
7 & GBM & human\_ensembl\_regulatory & 90.3 & 89.7 & 89.6 \\
8 & GBM & human\_nontata\_promoters & 92.7 & 93.5 & 92.9 \\
9 & GBM & human\_ocr\_ensembl & 74.0 & 74.7 & 73.6 \\
\midrule
10 & GUE & H3 & 85.0 & 85.1 & 83.3 \\
11 & GUE & H3K14ac & 69.6 & 69.3 & 67.2 \\
12 & GUE & H3K36me3 & 71.1 & 71.3 & 70.4 \\
13 & GUE & H3K4me1 & 68.7 & 69.5 & 68.3 \\
14 & GUE & H3K4me2 & 66.9 & 66.3 & 65.7 \\
15 & GUE & H3K4me3 & 62.9 & 62.2 & 63.0 \\
16 & GUE & H3K79me3 & 78.7 & 79.9 & 77.6 \\
17 & GUE & H3K9ac & 73.2 & 73.5 & 72.2 \\
18 & GUE & H4 & 86.6 & 86.3 & 85.6 \\
19 & GUE & H4ac & 67.1 & 67.2 & 66.8 \\
20 & GUE & mouse\_0 & 66.7 & 68.2 & 64.3 \\
21 & GUE & mouse\_1 & 88.6 & 88.8 & 88.0 \\
22 & GUE & mouse\_2 & 82.0 & 82.9 & 84.3 \\
23 & GUE & mouse\_3 & 72.1 & 84.2 & 77.0 \\
24 & GUE & mouse\_4 & 66.4 & 65.6 & 65.3 \\
25 & GUE & prom\_300\_all & 91.2 & 91.2 & 90.7 \\
26 & GUE & prom\_300\_notata & 94.5 & 95.4 & 94.9 \\
27 & GUE & prom\_300\_tata & 79.5 & 76.0 & 80.3 \\
28 & GUE & prom\_core\_all & 81.0 & 80.9 & 80.0 \\
29 & GUE & prom\_core\_notata & 82.4 & 82.7 & 82.3 \\
30 & GUE & prom\_core\_tata & 78.6 & 81.3 & 76.5 \\
31 & GUE & splice\_reconstructed & 82.3 & 80.8 & 80.9 \\
32 & GUE & tf\_0 & 83.4 & 82.2 & 81.3 \\
33 & GUE & tf\_1 & 83.6 & 84.4 & 82.9 \\
34 & GUE & tf\_2 & 77.1 & 77.3 & 73.6 \\
35 & GUE & tf\_3 & 71.2 & 72.0 & 68.8 \\
36 & GUE & tf\_4 & 82.4 & 83.0 & 81.3 \\
\midrule
37 & snATAC-seq & Human/Mouse brain & 35.4 & 35.9 & 34.4 \\
\midrule
38 & Multi-SCREEN & Human cCRE & 31.1 & 30.9 & 30.3 \\
\midrule
39 & NT & H2AFZ & 73.9 & 73.5 & 72.2 \\
40 & NT & H3K27ac & 73.4 & 72.1 & 71.4 \\
41 & NT & H3K27me3 & 80.2 & 80.7 & 80.7 \\
42 & NT & H3K36me3 & 79.5 & 78.3 & 78.6 \\
43 & NT & H3K4me1 & 74.4 & 74.5 & 73.8 \\
44 & NT & H3K4me2 & 78.6 & 78.7 & 77.6 \\
45 & NT & H3K4me3 & 83.3 & 83.2 & 82.6 \\
46 & NT & H3K9ac & 76.5 & 76.6 & 74.6 \\
47 & NT & H3K9me3 & 70.4 & 70.9 & 70.7 \\
48 & NT & H4K20me1 & 81.3 & 81.7 & 81.1 \\
49 & NT & enhancers & 74.1 & 73.9 & 73.2 \\
50 & NT & enhancers\_types & 70.5 & 70.3 & 69.7 \\
51 & NT & promoter\_all & 86.7 & 85.6 & 85.0 \\
52 & NT & promoter\_no\_tata & 87.1 & 87.1 & 87.0 \\
53 & NT & promoter\_tata & 84.3 & 82.0 & 84.1 \\
54 & NT & splice\_sites\_acceptors & 83.7 & 81.6 & 81.0 \\
55 & NT & splice\_sites\_all & 76.5 & 72.8 & 70.8 \\
56 & NT & splice\_sites\_donors & 83.5 & 80.6 & 78.7 \\
\bottomrule
\end{tabular}
\caption{Per-task Precision (\%) for ablation variants.}
\label{tab:per_task_prec_ablation}
\end{table}

\begin{table}[H]
\centering
\scriptsize
\setlength{\tabcolsep}{3pt}
\renewcommand{\arraystretch}{0.95}
\begin{tabular}{rllccc}
\toprule
\# & Benchmark & Task & DNAbert2 & NT & Grover \\
\midrule
1 & GBM & demo\_coding\_vs\_intergenomic\_seqs & 90.2 & 89.0 & 91.5 \\
2 & GBM & demo\_human\_or\_worm & 95.8 & 93.8 & 96.4 \\
3 & GBM & drosophila\_enhancers\_stark & 68.5 & 69.0 & 68.7 \\
4 & GBM & dummy\_mouse\_enhancers\_ensembl & 73.8 & 75.4 & 73.9 \\
5 & GBM & human\_enhancers\_cohn & 72.8 & 71.7 & 73.6 \\
6 & GBM & human\_enhancers\_ensembl & 89.3 & 83.3 & 90.5 \\
7 & GBM & human\_ensembl\_regulatory & 89.8 & 90.7 & 90.3 \\
8 & GBM & human\_nontata\_promoters & 93.3 & 85.6 & 95.5 \\
9 & GBM & human\_ocr\_ensembl & 73.8 & 72.0 & 77.7 \\
\midrule
10 & GUE & H3 & 85.5 & 82.9 & 87.4 \\
11 & GUE & H3K14ac & 68.7 & 64.8 & 70.5 \\
12 & GUE & H3K36me3 & 71.4 & 68.2 & 72.4 \\
13 & GUE & H3K4me1 & 68.7 & 66.7 & 68.9 \\
14 & GUE & H3K4me2 & 63.8 & 63.8 & 63.9 \\
15 & GUE & H3K4me3 & 63.1 & 58.1 & 65.0 \\
16 & GUE & H3K79me3 & 78.9 & 75.8 & 79.5 \\
17 & GUE & H3K9ac & 74.8 & 72.2 & 75.7 \\
18 & GUE & H4 & 87.8 & 83.3 & 85.8 \\
19 & GUE & H4ac & 67.0 & 63.4 & 69.8 \\
20 & GUE & mouse\_0 & 66.4 & 65.3 & 71.9 \\
21 & GUE & mouse\_1 & 89.4 & 85.9 & 90.2 \\
22 & GUE & mouse\_2 & 87.5 & 81.1 & 86.3 \\
23 & GUE & mouse\_3 & 74.9 & 72.4 & 82.8 \\
24 & GUE & mouse\_4 & 65.6 & 60.6 & 70.5 \\
25 & GUE & prom\_300\_all & 90.5 & 92.0 & 91.2 \\
26 & GUE & prom\_300\_notata & 95.1 & 94.6 & 94.9 \\
27 & GUE & prom\_300\_tata & 78.6 & 82.1 & 78.3 \\
28 & GUE & prom\_core\_all & 80.5 & 80.2 & 80.7 \\
29 & GUE & prom\_core\_notata & 82.2 & 81.2 & 82.8 \\
30 & GUE & prom\_core\_tata & 75.9 & 83.7 & 79.9 \\
31 & GUE & splice\_reconstructed & 84.0 & 85.9 & 87.6 \\
32 & GUE & tf\_0 & 81.7 & 80.3 & 83.2 \\
33 & GUE & tf\_1 & 84.5 & 81.9 & 83.7 \\
34 & GUE & tf\_2 & 77.5 & 72.7 & 79.0 \\
35 & GUE & tf\_3 & 68.2 & 67.1 & 72.2 \\
36 & GUE & tf\_4 & 84.0 & 81.4 & 85.8 \\
\midrule
37 & snATAC-seq & Human/Mouse brain & 30.4 & 30.9 & 33.7 \\
\midrule
38 & Multi-SCREEN & Human cCRE & 31.6 & 31.0 & 33.3 \\
\midrule
39 & NT & H2AFZ & 73.5 & 73.5 & 73.4 \\
40 & NT & H3K27ac & 73.1 & 71.0 & 73.3 \\
41 & NT & H3K27me3 & 80.0 & 80.0 & 80.4 \\
42 & NT & H3K36me3 & 80.1 & 78.5 & 80.1 \\
43 & NT & H3K4me1 & 73.7 & 73.0 & 73.9 \\
44 & NT & H3K4me2 & 78.2 & 77.4 & 78.2 \\
45 & NT & H3K4me3 & 83.1 & 82.8 & 83.6 \\
46 & NT & H3K9ac & 76.7 & 73.6 & 77.2 \\
47 & NT & H3K9me3 & 71.9 & 67.9 & 71.3 \\
48 & NT & H4K20me1 & 81.4 & 80.3 & 81.4 \\
49 & NT & enhancers & 73.8 & 74.5 & 75.0 \\
50 & NT & enhancers\_types & 50.7 & 48.9 & 50.0 \\
51 & NT & promoter\_all & 86.7 & 85.9 & 86.9 \\
52 & NT & promoter\_no\_tata & 86.8 & 87.7 & 87.3 \\
53 & NT & promoter\_tata & 85.8 & 86.4 & 86.0 \\
54 & NT & splice\_sites\_acceptors & 85.6 & 96.6 & 87.5 \\
55 & NT & splice\_sites\_all & 79.6 & 95.7 & 87.2 \\
56 & NT & splice\_sites\_donors & 84.0 & 97.5 & 91.4 \\
\bottomrule
\end{tabular}
\caption{Per-task Recall (\%) for tokenizer-strategy comparison.}
\label{tab:per_task_recall_tokenizer}
\end{table}

\begin{table}[H]
\centering
\scriptsize
\setlength{\tabcolsep}{3pt}
\renewcommand{\arraystretch}{0.95}
\begin{tabular}{rllccc}
\toprule
\# & Benchmark & Task & w/o Len & w/o Part & w/o Pri \\
\midrule
1 & GBM & demo\_coding\_vs\_intergenomic\_seqs & 90.4 & 90.6 & 90.1 \\
2 & GBM & demo\_human\_or\_worm & 95.2 & 95.8 & 95.6 \\
3 & GBM & drosophila\_enhancers\_stark & 76.2 & 74.6 & 71.0 \\
4 & GBM & dummy\_mouse\_enhancers\_ensembl & 73.1 & 73.3 & 79.6 \\
5 & GBM & human\_enhancers\_cohn & 74.3 & 73.2 & 73.6 \\
6 & GBM & human\_enhancers\_ensembl & 88.4 & 88.6 & 87.6 \\
7 & GBM & human\_ensembl\_regulatory & 89.4 & 88.8 & 88.8 \\
8 & GBM & human\_nontata\_promoters & 93.1 & 93.9 & 93.2 \\
9 & GBM & human\_ocr\_ensembl & 73.9 & 74.6 & 73.5 \\
\midrule
10 & GUE & H3 & 84.9 & 84.9 & 83.3 \\
11 & GUE & H3K14ac & 69.1 & 69.7 & 67.5 \\
12 & GUE & H3K36me3 & 71.2 & 71.5 & 70.3 \\
13 & GUE & H3K4me1 & 68.3 & 69.1 & 68.0 \\
14 & GUE & H3K4me2 & 64.1 & 63.1 & 62.7 \\
15 & GUE & H3K4me3 & 62.5 & 61.6 & 62.7 \\
16 & GUE & H3K79me3 & 78.7 & 79.9 & 77.7 \\
17 & GUE & H3K9ac & 73.2 & 73.7 & 72.5 \\
18 & GUE & H4 & 86.3 & 86.3 & 85.3 \\
19 & GUE & H4ac & 67.0 & 66.8 & 66.8 \\
20 & GUE & mouse\_0 & 66.7 & 68.0 & 64.3 \\
21 & GUE & mouse\_1 & 88.6 & 88.8 & 88.0 \\
22 & GUE & mouse\_2 & 82.0 & 82.9 & 84.1 \\
23 & GUE & mouse\_3 & 72.0 & 84.1 & 77.0 \\
24 & GUE & mouse\_4 & 66.1 & 65.5 & 65.1 \\
25 & GUE & prom\_300\_all & 91.2 & 91.2 & 90.7 \\
26 & GUE & prom\_300\_notata & 94.5 & 95.4 & 94.9 \\
27 & GUE & prom\_300\_tata & 79.3 & 75.8 & 80.2 \\
28 & GUE & prom\_core\_all & 80.9 & 80.9 & 80.0 \\
29 & GUE & prom\_core\_notata & 82.4 & 82.7 & 82.3 \\
30 & GUE & prom\_core\_tata & 77.9 & 80.8 & 76.5 \\
31 & GUE & splice\_reconstructed & 82.1 & 80.4 & 80.3 \\
32 & GUE & tf\_0 & 83.0 & 81.8 & 80.7 \\
33 & GUE & tf\_1 & 82.6 & 83.7 & 82.0 \\
34 & GUE & tf\_2 & 75.8 & 77.3 & 73.0 \\
35 & GUE & tf\_3 & 71.0 & 70.4 & 67.5 \\
36 & GUE & tf\_4 & 82.4 & 82.9 & 81.3 \\
\midrule
37 & snATAC-seq & Human/Mouse brain & 33.0 & 32.9 & 32.5 \\
\midrule
38 & Multi-SCREEN & Human cCRE & 31.2 & 31.3 & 30.9 \\
\midrule
39 & NT & H2AFZ & 73.8 & 72.2 & 72.1 \\
40 & NT & H3K27ac & 72.8 & 71.7 & 71.0 \\
41 & NT & H3K27me3 & 80.0 & 80.6 & 79.6 \\
42 & NT & H3K36me3 & 79.3 & 78.3 & 78.5 \\
43 & NT & H3K4me1 & 74.2 & 74.0 & 73.0 \\
44 & NT & H3K4me2 & 78.4 & 78.4 & 76.9 \\
45 & NT & H3K4me3 & 83.3 & 83.2 & 82.6 \\
46 & NT & H3K9ac & 76.4 & 76.5 & 74.3 \\
47 & NT & H3K9me3 & 70.3 & 70.7 & 70.7 \\
48 & NT & H4K20me1 & 81.3 & 81.6 & 81.1 \\
49 & NT & enhancers & 73.9 & 73.9 & 73.0 \\
50 & NT & enhancers\_types & 50.9 & 50.8 & 49.8 \\
51 & NT & promoter\_all & 86.3 & 85.6 & 85.0 \\
52 & NT & promoter\_no\_tata & 87.1 & 87.1 & 86.7 \\
53 & NT & promoter\_tata & 84.3 & 82.0 & 84.1 \\
54 & NT & splice\_sites\_acceptors & 83.7 & 81.5 & 81.0 \\
55 & NT & splice\_sites\_all & 76.4 & 72.8 & 70.2 \\
56 & NT & splice\_sites\_donors & 83.4 & 80.6 & 78.4 \\
\bottomrule
\end{tabular}
\caption{Per-task Recall (\%) for ablation variants.}
\label{tab:per_task_recall_ablation}
\end{table}

\begin{table}[H]
\centering
\scriptsize
\setlength{\tabcolsep}{3pt}
\renewcommand{\arraystretch}{0.95}
\begin{tabular}{rllccc}
\toprule
\# & Benchmark & Task & Base & EvoLen & $\Delta$ \\
\midrule
1 & GBM & demo\_coding\_vs\_intergenomic\_seqs & 82.1 & 81.8 & -0.2 \\
2 & GBM & demo\_human\_or\_worm & 92.3 & 92.7 & +0.4 \\
3 & GBM & drosophila\_enhancers\_stark & 54.3 & 56.4 & +2.1 \\
4 & GBM & dummy\_mouse\_enhancers\_ensembl & 53.8 & 64.2 & +10.4 \\
5 & GBM & human\_enhancers\_cohn & 45.3 & 50.1 & +4.8 \\
6 & GBM & human\_enhancers\_ensembl & 79.4 & 79.3 & -0.1 \\
7 & GBM & human\_ensembl\_regulatory & 84.2 & 84.2 & +0.0 \\
8 & GBM & human\_nontata\_promoters & 89.3 & 89.9 & +0.6 \\
9 & GBM & human\_ocr\_ensembl & 51.1 & 50.6 & -0.5 \\
\midrule
10 & GUE & H3 & 70.7 & 72.4 & +1.6 \\
11 & GUE & H3K14ac & 39.9 & 41.8 & +1.9 \\
12 & GUE & H3K36me3 & 43.7 & 45.1 & +1.4 \\
13 & GUE & H3K4me1 & 38.2 & 37.8 & -0.4 \\
14 & GUE & H3K4me2 & 29.2 & 30.6 & +1.5 \\
15 & GUE & H3K4me3 & 28.8 & 28.7 & -0.1 \\
16 & GUE & H3K79me3 & 57.1 & 59.9 & +2.7 \\
17 & GUE & H3K9ac & 47.6 & 50.4 & +2.7 \\
18 & GUE & H4 & 74.4 & 75.6 & +1.1 \\
19 & GUE & H4ac & 39.4 & 36.7 & -2.7 \\
20 & GUE & mouse\_0 & 33.9 & 45.7 & +11.8 \\
21 & GUE & mouse\_1 & 79.3 & 79.4 & +0.1 \\
22 & GUE & mouse\_2 & 72.6 & 73.8 & +1.2 \\
23 & GUE & mouse\_3 & 54.8 & 59.5 & +4.7 \\
24 & GUE & mouse\_4 & 32.1 & 35.9 & +3.8 \\
25 & GUE & prom\_300\_all & 83.5 & 83.0 & -0.5 \\
26 & GUE & prom\_300\_notata & 89.8 & 90.8 & +1.0 \\
27 & GUE & prom\_300\_tata & 59.5 & 60.5 & +1.0 \\
28 & GUE & prom\_core\_all & 63.6 & 63.1 & -0.5 \\
29 & GUE & prom\_core\_notata & 64.5 & 66.0 & +1.5 \\
30 & GUE & prom\_core\_tata & 66.4 & 64.1 & -2.3 \\
31 & GUE & splice\_reconstructed & 75.5 & 75.6 & +0.1 \\
32 & GUE & tf\_0 & 65.1 & 65.8 & +0.7 \\
33 & GUE & tf\_1 & 67.7 & 68.3 & +0.6 \\
34 & GUE & tf\_2 & 53.1 & 57.0 & +3.9 \\
35 & GUE & tf\_3 & 45.1 & 46.1 & +1.0 \\
36 & GUE & tf\_4 & 62.1 & 68.8 & +6.7 \\
\midrule
37 & snATAC-seq & Human/Mouse brain & 10.3 & 13.7 & +3.4 \\
\midrule
38 & Multi-SCREEN & Human cCRE & 22.7 & 22.5 & -0.2 \\
\midrule
39 & NT & H2AFZ & 47.9 & 49.1 & +1.2 \\
40 & NT & H3K27ac & 46.5 & 46.0 & -0.5 \\
41 & NT & H3K27me3 & 57.5 & 58.8 & +1.3 \\
42 & NT & H3K36me3 & 60.5 & 61.7 & +1.3 \\
43 & NT & H3K4me1 & 45.6 & 47.9 & +2.3 \\
44 & NT & H3K4me2 & 54.3 & 58.5 & +4.2 \\
45 & NT & H3K4me3 & 64.0 & 66.1 & +2.2 \\
46 & NT & H3K9ac & 52.2 & 54.6 & +2.4 \\
47 & NT & H3K9me3 & 39.4 & 42.0 & +2.6 \\
48 & NT & H4K20me1 & 66.9 & 69.4 & +2.5 \\
49 & NT & enhancers & 51.1 & 53.5 & +2.4 \\
50 & NT & enhancers\_types & 48.3 & 46.5 & -1.8 \\
51 & NT & promoter\_all & 73.1 & 74.9 & +1.8 \\
52 & NT & promoter\_no\_tata & 76.5 & 77.0 & +0.5 \\
53 & NT & promoter\_tata & 75.0 & 76.9 & +1.9 \\
54 & NT & splice\_sites\_acceptors & 63.9 & 68.7 & +4.8 \\
55 & NT & splice\_sites\_all & 69.2 & 68.3 & -0.8 \\
56 & NT & splice\_sites\_donors & 73.9 & 68.1 & -5.8 \\
\bottomrule
\end{tabular}
\caption{Per-task MCC (\%) for the compute-matched comparison at 200k pretraining steps. Both models use identical fine-tuning hyperparameters per task; the only difference is the tokenizer. EvoLen improves on 43 of 56 tasks.}
\label{tab:per_task_mcc_200k}
\end{table}

\subsection{Stratification Window Size}\label{sec:window_size}

The 100~bp stratification window balances two competing pressures: shorter windows track base-level phyloP more faithfully but fragment motifs, while longer windows smooth away the conservation signal the partition is meant to capture. Table~\ref{tab:S11_window} reports chromosome~1 diagnostics across window sizes. \textit{SD} is the standard deviation of the window-mean phyloP; \textit{Lag-1} is the autocorrelation between adjacent windows; \textit{Agreement} and $\kappa$ compare base-level conservation labels against the 100~bp reference partition. Dispersion falls monotonically with window size, lag-1 autocorrelation peaks near 100~bp, and label stability is highest at 50--200~bp. The lower panel reports the probability that a motif of a given length is split across a bin boundary: at 100~bp, 89--95\% of 6--12~bp motifs fall entirely within a single bin, whereas motif-scale windows split most of them.

\begin{table}[htbp]
\centering
\small
\caption{PhyloP window-size diagnostics on chromosome~1 (upper) and motif boundary-crossing probability by bin size (lower).}\label{tab:S11_window}
\setlength{\tabcolsep}{3.5pt}
\begin{tabular}{@{}rrrrrr@{}}
\toprule
Window (bp) & Windows & SD & Lag-1 & Agreement\% & $\kappa$ \\
\midrule
5 & 49,791,284 & 0.360 & 0.256 & 89.12 & 0.293 \\
10 & 24,895,642 & 0.285 & 0.363 & 90.74 & 0.399 \\
25 & 9,958,256 & 0.224 & 0.521 & 93.45 & 0.562 \\
50 & 4,979,128 & 0.195 & 0.608 & 95.67 & 0.703 \\
100 & 2,489,564 & \textbf{0.175} & \textbf{0.614} & --- & --- \\
200 & 1,244,782 & 0.157 & 0.569 & 95.70 & 0.699 \\
500 & 497,912 & 0.134 & 0.511 & 93.18 & 0.525 \\
\bottomrule
\end{tabular}

\vspace{0.5em}

\small
\begin{tabular}{@{}rccc@{}}
\toprule
Bin (bp) & 6~bp motif & 8~bp motif & 12~bp motif \\
\midrule
10 & 0.500 & 0.700 & 1.000 \\
25 & 0.200 & 0.280 & 0.440 \\
50 & 0.100 & 0.140 & 0.220 \\
100 & \textbf{0.050} & \textbf{0.070} & \textbf{0.110} \\
200 & 0.025 & 0.035 & 0.055 \\
500 & 0.010 & 0.014 & 0.022 \\
\bottomrule
\end{tabular}
\end{table}

\subsection{Category-Specific Vocabulary Distinctness}\label{sec:pool_distinctness}

The three category-specific BPE vocabularies are not redundant. Table~\ref{tab:S12_pool} reports pairwise Jaccard overlap and per-pool compression statistics. \textit{bp/Tok} is the mean number of base pairs per token; $H$ is the unigram Shannon entropy in bits; and $H_{\mathrm{norm}}$ is $H/\log_2|V|$. Conserved and Accelerated vocabularies share only 32--34\% of their tokens at every vocabulary size, while Conserved and Neutral overlap substantially more (52--58\%), consistent with conservation forming a gradient rather than discrete classes. Normalised entropy is consistently highest for the Neutral pool and lowest for Accelerated, a gap of roughly 0.028 that is stable across vocabulary sizes, indicating that the Accelerated pool concentrates usage on fewer tokens.

\begin{table}[htbp]
\centering
\small
\caption{Pairwise vocabulary Jaccard overlap (upper) and per-pool compression and entropy (lower) for the three category-specific BPEs.}\label{tab:S12_pool}
\begin{tabular}{@{}lccc@{}}
\toprule
Vocab & Con--Neu & Acc--Con & Acc--Neu \\
\midrule
2048 & 0.579 & 0.324 & 0.389 \\
3072 & 0.523 & 0.325 & 0.359 \\
4096 & 0.547 & 0.330 & 0.364 \\
5120 & 0.564 & 0.336 & 0.375 \\
\bottomrule
\end{tabular}

\vspace{0.5em}

\small
\setlength{\tabcolsep}{3.5pt}
\begin{tabular}{@{}llccc@{}}
\toprule
Vocab & Pool & bp/Tok & $H$ & $H_{\mathrm{norm}}$ \\
\midrule
2048 & Conserved & 4.747 & 9.484 & 0.863 \\
2048 & Neutral & 4.829 & 9.541 & 0.868 \\
2048 & Accelerated & 4.677 & 9.254 & 0.842 \\
\midrule
3072 & Conserved & 4.987 & 9.947 & 0.859 \\
3072 & Neutral & 5.083 & 10.002 & 0.864 \\
3072 & Accelerated & 4.912 & 9.686 & 0.836 \\
\midrule
4096 & Conserved & 5.152 & 10.262 & 0.855 \\
4096 & Neutral & 5.265 & 10.332 & 0.861 \\
4096 & Accelerated & 5.081 & 9.997 & 0.833 \\
\midrule
5120 & Conserved & 5.279 & 10.503 & 0.853 \\
5120 & Neutral & 5.408 & 10.586 & 0.859 \\
5120 & Accelerated & 5.213 & 10.240 & 0.831 \\
\bottomrule
\end{tabular}
\end{table}

\end{document}